\documentclass[preprint,12pt]{elsarticle}

\usepackage{lmodern}
\usepackage{amssymb}
\usepackage{amsmath}
\usepackage[figuresright]{rotating}
\usepackage{hyperref}
\usepackage{enumitem}
\usepackage{graphicx}
\usepackage{tikz}
\usepackage{colortbl}
\usepackage{xcolor}
\usepackage{subfig}
\usepackage{verbatim}
\usepackage{comment}
\usepackage{booktabs}
\usepackage{placeins}

\begin{document}

\begin{frontmatter}

\title{LLM-based Models for Detecting Emerging Topics in Service Feedback}

\author[1]{Mahsa Tavakoli}
\author[2]{Ruth Bankey}
\author[1]{Cristián Bravo}

\address[1]{Department of Statistical and Actuarial Sciences, The University of Western Ontario, London, Ontario, Canada. 
\{mtavako5, cbravoro\}@uwo.ca}

\address[2]{Canada Revenue Agency. Ruth.Bankey@cra-arc.gc.ca}

\begin{abstract}
Enhancing the analysis of service feedback data quality is crucial for public sector organizations, particularly tax administrations, where trust and compliance depend on fair and effective service delivery. 
A key challenge in this domain is to ensure that service quality remains consistent and inclusive across a diverse population, particularly as feedback volumes increase, a dynamic that can reveal disparities in how services are experienced, a topic of growing importance in public service analytics aimed at promoting equitable delivery.

Traditional methods for analyzing service feedback and supporting the detection of potential disparities or bias often rely on manual processes or static expert-driven indicators, which are not easily scalable and may struggle to capture the nuanced patterns present in textual data. In contrast, data-informed approaches that take advantage of actual feedback offer a more robust and dynamic means of identifying emerging service quality issues and promoting equitable service delivery.

This paper presents a novel methodology that integrates LLMs, statistical techniques, and human-AI teaming to improve multilingual customer feedback analysis, with the primary goal of detecting emerging topics in service quality, areas that can also indicate potential biases. Using AI-powered tools alongside expert oversight, our aim is to identify service delivery trends and support equitable, responsive outcomes across diverse demographic groups.
Our approach incorporates fine-tuned and quantized LLMs optimized for accuracy while minimizing computational demands. Validation was performed using similarity analysis and an evaluation survey by tax officers with direct experience and expertise in tax service feedback operations, demonstrating improved alignment with expert assessments compared to baseline models.%%%

%Although timeliness remains a key dimension of service evaluation, our approach highlights additional emerging concerns across demographic groups, helping to broaden the scope of service quality analysis in multilingual feedback.%%%
This study addresses the challenges of adapting LLMs to specific organizational contexts through a targeted fine-tuning approach. By integrating a human-in-the-loop methodology, we mitigate LLM fabrication and ensure reliable, context-aware outputs. The results highlight the practicality and effectiveness of this methodology in improving service quality, responsiveness, and decision-making in public sector operations. This research contributes to the development of responsible AI systems that prioritize fairness, inclusivity, and public trust in automated service delivery.
\end{abstract}

\begin{keyword}
Multilingual Feedback Analysis, Customer Service, Large Language Models (LLMs), Trend Detection, Fine-tuning, Quantization, Fairness,  Human-in-the-Loop Evaluation, Topic Categorization.
\end{keyword}

\end{frontmatter}

\section{Introduction} \label{sec:Intro}

Enhancing customer service quality has become a cornerstone of modern organizational success, as it directly influences customer satisfaction, loyalty, and trust \cite{scott2024revealing}. In public sector services, such as tax administrations, the stakes are even higher. An efficient and responsive service ensures compliance and fosters public satisfaction, which is critical to maintaining trust in government institutions. However, public satisfaction with tax services in Canada and the United States has been moderate, with room for improvement. In Canada, the overall Client Satisfaction Index (CSI) score was 63 out of 100, reflecting a moderately high level of satisfaction \cite{citizenfirst2020}. 
In the United States, citizen satisfaction with federal government services reached 69.7 out of 100 in 2024, the highest since 2017 \cite{fedscoop2024}.
These figures underscore the need for public organizations to ensure fairness in treating diverse demographic groups and responding promptly and effectively to customer needs.
%%%
One way to assess fairness is to analyze the topics that emerge more frequently in feedback across different demographic segments. Significant differences in topic frequency can indicate disparities in service experience, an important signal of possible bias. Although this paper does not aim to detect bias directly, identifying such patterns supports fairness objectives and aligns with the broader definition of bias as uneven treatment or outcomes among groups. Addressing these disparities, whether they stem from unconscious individual biases or systemic issues in automated systems, is essential to maintain public trust and ensure equitable service delivery.%%%
%However, achieving this requires confronting unconscious biases in individuals and explicit biases in automated systems, both of which can lead to unequal treatment. Such disparities risk undermining public trust and compromising equitable service delivery, making fairness and inclusivity essential priorities for organizational decision making \cite{scott2024revealing}.
Despite this critical need, existing approaches in public service organizations often rely on manual processes or quantitative indicators that are not data-driven, which are insufficient to capture the complexity of taxpayer concerns or address the nuanced bias and inequality patterns that arise in diverse demographic groups\cite{Michael2024}. 
Manual methods lack the efficiency and scope required to systematically detect and address biases across diverse demographic groups, especially as the volume and diversity of feedback increases. These processes are often slow, resource intensive and prone to human error and intrinsic variability, making it difficult to identify patterns or trends in a timely manner. This issue can be exacerbated in countries with more than one official language. This limitation becomes particularly critical when analyzing feedback at scale, where nuanced insights, such as disparities in service delivery between demographic segments, may be missed. 
In addition, automating real-time service quality evaluations based on customer behavior has demonstrated notable improvements in user engagement, evaluation accuracy, and overall efficiency of service operations \cite{Li2023, chen2025exploring}. These advances ensure that organizations not only address underlying biases, but also streamline processes, offering personalized and equitable services at scale, which further enhances customer satisfaction and operational performance.

To achieve this goal, the integration of advanced tools like artificial intelligence, particularly when paired with statistical techniques, can be highly effective. These tools can analyze large datasets, detect inequality patterns, and ensure that decision-making processes are fair and inclusive on a large scale. Recognizing this need, we aim to develop a methodology that integrates LLMs, statistical techniques, and human expertise. 

%%%%%%%%%%%%%%%%%%current research
Previous research has increasingly emphasized the importance of identifying disparities in service experience and bias, given its profound implications for fairness and equality in various social sectors \cite{gunarathne2022racial}. This growing focus stems from the need to develop more equitable systems and data-informed decision-making processes. Numerous studies have underscored the importance of addressing bias\cite{guilbeault2024online,zheng2024does}, demonstrating how targeted interventions can foster inclusivity and improve the overall effectiveness of systems that influence both daily life and institutional operations. By engaging in a comprehensive investigation of disparities, scholars contribute to the development of a more just and equitable society \cite{gunarathne2022racial}.
%%%

Recent research increasingly emphasizes detecting emerging trends in customer feedback across demographic groups as a foundation for assessing fairness. While our study does not directly detect bias, identifying consistent group-level differences in concerns can signal potential inequalities. Common methodologies include statistical techniques such as factor analysis, ANOVA, and regression to examine disparities that may reflect systemic service gaps or unintended biases. For instance, Linzmajer et al. \cite{linzmajer2020customer} employed lab and field experiments to assess how identity factors influence customer perceptions in service encounters.
With advances in artificial intelligence (AI) and machine learning (ML), researchers have increasingly turned to more sophisticated computational techniques to support the detection of disparities and promote fairness. %This shift has been particularly pronounced in the analysis of text and language data, where deep learning methods excel due to their ability to process complex sequential patterns. 

%%%
In addition to general language processing tasks, researchers have increasingly turned their attention to analyzing customer feedback and reviews to uncover disparities in how service quality is experienced in different demographic groups \cite{xie2024self}. Since customer reviews play a vital role in shaping perceptions of service performance, identifying patterns in topic prevalence by demographic segment can support fairer evaluations and help organizations address unequal service outcomes \cite{guo2020positive}. These emerging topic trends offer an indirect but meaningful way to surface potential disparities that may reflect underlying biases. Studies have confirmed that feedback and reviews significantly influence operational decisions, making them a valuable area for equity-focused analysis \cite{chen2021reviews}.% For example, Vollaard et al. \cite{vollaard2022bias} identified unintended disparities in texts that reference identity-related terms, while Asr et al. \cite{asr2021gender} highlighted gender-based imbalances in abusive language detection systems.

With the growing use of textual data in customer experience evaluation, NLP techniques have become crucial for large-scale analysis.% Transfer learning approaches, such as those by Mozafari et al. \cite{mozafari2020hate}, use pre-trained models like BERT to detect problematic content and mitigate bias during fine-tuning. 
More recently, LLMs have been employed to model topic distributions and explore contextual bias, as demonstrated by Kumar et al. \cite{kumar2024bias}. While prior work emphasizes explicit bias detection, our study focuses on emerging multilingual feedback patterns that, when analyzed by demographic group, may highlight disparities in public service delivery.
While prior research has advanced automated bias detection in customer service, aligning methodologies with an organization's specific goals and context remains a challenge. Existing statistical approaches often overlook the nuanced realities of service delivery. Detecting emerging feedback topics by demographic group offers a more scalable and context-sensitive method for uncovering disparities. Although not explicitly labeled as bias, such patterns can highlight unequal experiences or unmet needs—key insights for enhancing fairness and service quality in public institutions.
Much of the research on text-based analysis remains monolingual, overlooking the multilingual needs of organizations serving diverse populations. Public sector entities, in particular, require tools that can detect biases and trends across languages to enhance accessibility and inclusivity \cite{Ravfogel2024}.

Although advanced NLP and LLM-based methods offer advantages in speed and accuracy, their application to bias detection remains limited. Many studies rely on general-purpose pre-trained models, which often fail to capture domain-specific nuances \cite{Hasan2022}. Fine-tuned models tailored to specific sectors, such as customer service and finance, have shown improved accuracy, yet their adoption is constrained by computational demands \cite{ Zhang2024}.
To address this, developing quantized, resource-efficient models is vital for broader accessibility. Moreover, integrating human expertise into the evaluation process is essential, as automated systems may miss subtle contextual cues. A hybrid human-AI approach improves fairness and reliability by combining large-scale automation with expert oversight \cite{pillai2024adoption}.

%%%%%%%%%%%%%%%%%%%%%%%%% General solution and challenges

AI-powered tools, particularly NLP, hold great promise for analyzing large-scale customer feedback and identifying emerging trends across demographic groups. These trends can reveal disparities in service perception or delivery \cite{Adam2021,Shah2023}. Miraz et al. \cite{Miraz2024} emphasized the effectiveness of AI chatbots in enhancing customer service through improved communication, usability, and user trust, with broad impacts on operations and engagement.
However, the deployment of such technologies must address algorithmic bias—systemic unfairness arising from biased data or design. To mitigate these risks, ethically aligned AI principles are essential, promoting fairness, transparency, and accountability in AI-driven applications \cite{Mogaji2022}.

Ethically aligned AI principles serve as a foundation for responsible AI development, emphasizing key attributes such as transparency, accountability, explainability, and fairness\cite{camilleri2024artificial}. \textbf{Fairness} ensures that AI systems provide equitable treatment to all demographic groups, preventing discrimination, and fostering inclusivity. \textbf{Accountability} introduces human oversight into the AI decision making process, allowing for the validation and refinement of AI-generated outcomes. \textbf{Transparency} plays a vital role in ensuring that AI processes remain open and comprehensible, allowing users to understand how decisions are made. Additionally, \textbf{explainability} enhances trust by providing clear, justifiable reasoning behind AI decisions, ensuring that users and stakeholders can interpret AI-driven conclusions with confidence.
By integrating these principles, organizations can develop AI systems that perform effectively and uphold ethical standards. This approach helps mitigate biases, fosters public trust, and ensures that AI technologies serve society in a fair and responsible manner.

%%%%%%%Solution:
To address existing gaps, our methodology uses fine-tuned LLMs to classify customer feedback into predefined categories (Service Quality Elements, SQEs) and analyze trends—emerging, persistent, or disappearing—across demographic groups. By focusing on Equity, Diversity, and Inclusion (EDI) and organizational goals, the tool captures issues often missed in traditional feedback systems. The approach blends statistical modeling with quantized LLMs for conceptual relevance and resource efficiency. Compared to baseline models, our LLMs showed stronger alignment with expert judgments, as confirmed through similarity analysis and expert surveys.

%To implement an AI-based tool that addresses the identified gaps, our proposed methodology leverages advanced LLMs to first classify a given feedback into pre-established groups/categories by the tax administration (in this case service quality elements, SQE), and then, by modeling the behavior of each SQE, detect emerging, persistent, and disappearing trends across diverse demographic groups. The focus is on identifying the underlying issues within customer feedback and accurately categorizing them into pre-defined categories. In doing so, the system ensures that patterns significant to particular groups – defined by Equity, Diversity, and Inclusion (EDI) criteria or aligned with the organization's strategic goals –are identified, even when such issues can be overlooked in traditional one-on-one feedback management processes. Our approach combines statistical modeling with fine-tuned and quantized LLMs, which are specifically tailored for optimal alignment with organizational processes. This ensures that the models are conceptually relevant not only to the organization but also efficient in resource utilization. In a comparative analysis with pre-trained baseline models, our fine-tuned LLMs demonstrated a higher level of agreement with expert judgments on topic categorization. These results were validated using similarity analysis and human evaluation surveys conducted by organizational experts, strengthening the effectiveness of this customized methodology.

Based on the objectives described, this study seeks to explore the following key research questions.

\begin{enumerate}
    \item How can we develop an automated human-in-the-loop system for thematic analysis of taxpayer feedback that ensures transparency and explainability while enabling fair and inclusive service evaluation across diverse demographic groups?%%%
    %How can we develop an automated system for thematic analysis of taxpayer feedback that effectively identifies bias while maintaining human oversight to ensure transparency and explainability?     
    \item How can we design a model that accurately identifies emerging patterns across different demographic groups within the feedback? 
    \item How can we implement automated topic modeling for feedback analysis to categorize responses into specific predefined categories?
\end{enumerate}
	
Our methodology is designed to address three critical challenges that often arise when deploying LLMs for operational improvements. One of the primary concerns is adapting LLMs to specific organizational contexts. Many organizations use domain-specific language or specialized terminology that differs significantly from the everyday language. Standard pre-trained LLMs may struggle to interpret or process this specialized vocabulary accurately, leading to potential miscommunication or inefficiencies. To overcome this, our approach incorporates a fine-tuning process in which the LLM is trained on organizational datasets. By doing so, the model gains a more profound understanding of industry-specific terminology, ensuring that it can effectively navigate the nuances of the organization's language.

Another major challenge involves mitigating confabulation, instances where LLMs generate inaccurate or fabricated outputs. Given the potential risks of misinformation, we adopt a human-in-the-loop approach to enhance reliability. In this framework, specialized officers actively review, validate, and refine the LLM’s results before they are applied in decision-making. This oversight ensures that the output generated is not only accurate, but also contextually relevant and actionable, thereby strengthening trust in the recommendations of the AI system. 
Finally, our methodology uses the efficiencies of LLMs to enhance process improvements, particularly in the area of feedback management. Traditional methods of categorizing and analyzing feedback can be time-consuming and resource-intensive. By integrating AI-driven automation, we streamline the feedback categorization process and facilitate trend detection with greater speed and precision. This shift reduces the manual workload, accelerates the generation of insights, and ultimately improves decision making throughout the organization. Through these strategies, our approach ensures that LLMs are not only effectively integrated into operational workflows but also optimized to meet the specific needs and challenges of the organization.
	
The paper is methodically structured to explore and validate the proposed multilingual LLM-based model to analyze service feedback in the Tax Service\footnote{The term “Tax Service” is used to anonymize the name of the tax authority and its country of origin.}
. It begins with an introduction and a comprehensive review of the literature, covering existing customer service improvement systems, the background on topic modeling, and identifying current research gaps. The methodology section details the processes of data collection and preprocessing, the general framework of the proposed system, and the specific techniques used for topic modeling and trend analysis. The Results section presents an in-depth analysis, including text visualizations, findings from topic modeling, model evaluation, and trend detection. Finally, the article concludes with a discussion of the key findings, acknowledges the limitations of the study, and suggests directions for future research to improve the scalability and adaptability of the model.

\section{Literature Review} \label{Literature Review}

\subsection{Background on Topic Modeling}

As we use topic modeling in our model as part of automated service feedback analysis and trend detection, our aim is to provide a literature review on well-known topic modeling techniques, including LDA-like methods\cite{zimmermann2024approaches}, BERT-based models and LLM-based frameworks\cite{li2023novel}. Topic modeling has experienced significant advances over the years, incorporating diverse methodologies to address evolving analytical challenges. 
Topic modeling has experienced significant advances over the years, incorporating diverse methodologies to address evolving analytical challenges. As reviewed by Murshed et al.\cite{Murshed2023} foundational approaches such as Latent Dirichlet Allocation (LDA), and Latent Semantic Analysis (LSA) have been pivotal in text analysis. LLMs have progressively augmented these traditional techniques to tackle challenges such as short text processing, event detection, and sentiment analysis, reflecting their growing relevance in contemporary applications.
%LDA, highlighted by Chaudhuri et al. \cite{Chaudhuri2024} remains a cornerstone for exploring thematic structures in texts, offering effective classification and broader applicability within natural language processing frameworks. Comparative analyses of these foundational methods demonstrate their unique advantages. LSA and NMF are valued for their interpretability and effectiveness in reducing dimensionality, while CTM enhances topic modeling by integrating correlations between topics\cite{Chauhan2022TopicMU}. Inaam ul Haq and Li (2023) further extend this trajectory by demonstrating how contextual modeling integrates traditional methods such as LSA and NMF with innovations in CTM \cite{Haq2023}.

Recent progress in topic modeling has also seen the integration of transformer-based models such as BERT, which enhances topic coherence and interpretability. Li et al. introduced a hybrid model that combines BERT with probabilistic topic modeling techniques to refine topic representations by using semantic word embeddings and multimodal supervision through labels and visual features \cite{li2023novel}. Rogers et al. provided an exhaustive review of the BERT architecture, underscoring its adaptability and effectiveness in various NLP tasks \cite{Rogers2020}. Mishra et al. demonstrated the efficacy of BERTopic in capturing evolving themes within computational economics, further highlighting the superiority of modern methods over traditional techniques \cite{Mishra2024}. 
Recent advances in the application of LLMs across diverse fields offered valuable insight into improving topic-modeling methodologies. Zhao et al. investigated the integration of LLMs into complex design tasks, addressing challenges such as fine-tuning and adapting models to specific real-world contexts, which were critical to effective topic modeling \cite{Zhao2024}. Sufi et al. focused on abstractive summarization and examined how LLM addressed semantic inconsistencies, offering implications for improving coherence and accuracy in topic detection \cite{Sufi2024}.
Applications beyond traditional domains also provided transferable insights. Tzelves et al. highlighted the role of LLM in surgical innovation, emphasizing methodological refinements that could be applied to topical modeling frameworks \cite{Tzelves2024}. Similarly, educational applications of LLM focus on improving semantic precision and contextual relevance crucial for interpreting complex feedback in educational settings, enhancing both understanding and participation in topic modeling \cite{Friha2024LLM}. Together, these studies underscored the versatility and evolving capabilities of LLMs, demonstrating their transformative potential to refine topic modeling techniques in various domains.

\textbf{Identified Gaps Emerging from the Literature}: Existing topic modeling studies often focus on broad topic extraction rather than expert-defined themes, limiting their applicability in domains requiring precise categorization. This misalignment hinders integration with operational processes. Moreover, many approaches rely on generic LLMs without fine-tuning them for domain-specific feedback, reducing their effectiveness in interpreting nuanced comments.
Our research addresses these gaps by exploring the benefits of fine-tuning LLMs on service manuals and applying quantization to enhance deployment efficiency in resource-constrained environments. These advancements support a transition from statistical to deep learning approaches and inform our development of a human-in-the-loop pipeline for accurate, fair, and scalable multilingual topic categorization in public service feedback.

%\textbf{Identified Gaps Emerging from the Literature}: A notable gap is the lack of targeted analysis based on expert-defined topics. Most existing topic modeling research emphasizes broad topic identification rather than extracting and analyzing text according to predefined themes. This limitation prevents models from aligning closely with stablished operational processes that have carefully designed such themes, and hinders their effectiveness in domain-specific applications where precise categorization is essential. Furthermore, existing studies predominantly employ LLMs for generic topic modeling without fine-tuning them for domain-specific feedback. Customizing LLMs to contexts by training them in specialized text significantly enhances their ability to accurately interpret nuanced taxpayer comments. Furthermore, deploying these models efficiently in resource-constrained environments requires quantization to optimize computational performance. Our research explores both the impact of fine-tuning models on service manuals and the efficiencies gained through quantization, assessing their broader organizational implications. Together, these studies emphasize the evolution of topic modeling from statistical to deep learning paradigms and inform our methodological choices for integrating fine-tuned and quantized LLMs into a human-in-the-loop pipeline for precise, fair and scalable topic categorization in multilingual public service feedback.

\subsection{Leveraging Machine Learning for Customer Feedback and Service Optimization}
As part of our methodology for analyzing taxpayer feedback at the Tax Service, we draw from the broader literature on machine learning and natural language processing applications in customer feedback analysis. %Although much of the research is aimed at commercial domains such as banking, retail, and e-Commerce, the foundational concepts, methods, and insights are transferable to the public sector. In this context, taxpayers are analogous to customers: both groups provide service-related feedback that organizations aim to understand and act upon. Therefore, this review synthesizes the literature on customer feedback optimization to inform the design of scalable, fair, and context-aware feedback analysis systems for Tax Service operations.

ML has emerged as a transformative tool for improving customer service\cite{pereira2022customer}, predicting customer behaviors\cite{schetgen2021predicting}, and leveraging customer feedback\cite{ojo2024prioritising} in various industries. Recent studies demonstrate the growing importance of ML models over traditional statistical methods in addressing customer-centric problems such as customer satisfaction, return behavior, and targeted marketing\cite{de2024hybrid}. 
Yi and Liu (2020)\cite{Yi2020} applied ML algorithms for customer sentiment analysis to recommend products and stores based on customer reviews. The authors demonstrated that ML techniques significantly outperform existing approaches, achieving high accuracy (98\%) in predicting product recommendations.% Similarly, Majumder et al. (2023) analyzed the perceived usefulness of online customer reviews (OCR) using text mining and sentiment analysis. They found that peripheral review features, such as star ratings and review length, significantly influenced review usefulness for search goods, offering actionable strategies to improve customer decision making. 
Hwang et al. (2020)\cite{Hwang2020} extended this line of inquiry by predicting customer return visits in airline services through ML classifiers. Their results achieved an accuracy of 83.42\%, highlighting the role of sentimental features in improving predictive performance. The study also identified that higher word counts in feedback enhanced prediction accuracy, showcasing the importance of review content quality in analyzing customer loyalty, and highlighting the growing reliance on ML to automate feedback processing.
In addition, while ML models are widely used, studies have compared their performance with traditional statistical methods\cite{maibaum2024selecting}.
%Oussama et al. (2024)\cite{Gafrej2024} evaluated artificial neural networks (ANN) and multiple linear regression (MLR) in predicting bank deposits. ANN consistently outperformed MLR, as evidenced by lower error rates, underscoring the superiority of ML in forecasting accuracy. 
%Similarly, Chou et al. (2022)\cite{Chou2022} explored the integration of Buy-Until-You-Die (BTYD) models with ML techniques to predict customer repurchase frequency. Combining BTYD output with Lasso regression significantly improved prediction accuracy, outperforming standalone methods and advanced neural networks. This integration highlights the complementarity of ML with traditional models.

However, ML models face challenges when applied to real-world data. Simester et al. (2020) \cite{Simester2020} examined the robustness of ML methods based on model-driven and classification in customer targeting. Although model-driven methods excelled under ideal conditions, their performance diminished under data challenges, such as covariate shift and imbalanced data. The classification methods performed poorly, prompting the need for further optimization in the handling of degraded datasets. These findings stress the importance of data quality in ML applications for customer service. In contrast, Zaghloul et al. (2024) \cite{Zaghloul2024} demonstrated that traditional ML models, such as Random Forest, outperformed deep learning approaches in predicting customer satisfaction in e-Commerce, achieving 92\% accuracy. The study identified critical satisfaction drivers, such as delivery time and order accuracy, and emphasized the practical value of simpler, interpretable ML methods over complex deep learning models.
The integration of ML with optimization-based models has shown substantial success in enhancing customer outcomes. Feldman et al. (2022) \cite{Feldman2022} compared a multinomial logit (MNL)-based model with ML algorithms to determine optimal product displays on Alibaba's marketplaces. The MNL-based approach outperformed the Alibaba ML model, generating a 28\% revenue increase and highlighting the potential of combining choice models with ML features for revenue optimization.

%##########################################################

NLP has become a crucial tool to improve customer service by automating feedback analysis, query resolution, and sentiment evaluation. Recent research highlights diverse applications of NLP techniques, offering insights into improving customer satisfaction, loyalty, and decision-making across various domains. %Piris and Gay \cite{Piris2021} leveraged NLP to analyze customer satisfaction from 12,000 customer returns, identifying eight themes influencing satisfaction levels. Their findings revealed that satisfaction arises from varying theme combinations, challenging traditional linear models of feedback analysis. This research laid the foundation for personalized and automated feedback processing, offering a deeper understanding of customer experiences.

Focusing on sentiment analysis (SA) in underserved languages, Islam et al. \cite{islam2025words} analyzed YouTube comments to explore public opinion about the war. They used a sentiment analysis tool along with an unsupervised BERT model to uncover key topics associated with the war.
Nair et al.\cite{Nair2025} proposed an NLP-driven approach to improve chatbot-based customer service by enabling natural and human-like interactions. Their model leverages sentiment analysis, entity extraction, and intent detection to improve response accuracy and customer satisfaction. Despite scalability and efficiency, challenges such as ambiguous language and the need for large training datasets persist. This aligns with our work in that both approaches aim to improve automated service interactions using NLP techniques.

For social network applications, sentiment analysis plays a key role in opinion mining. \cite{krosuri2023novel} introduced a hybrid model that integrates ResNeXt with a recurrent neural framework to improve multiclass classification. This method improves accuracy by removing noise through unsupervised processing and minimizing annotation efforts compared to traditional techniques. The model was tested on Amazon and Twitter datasets.
To improve customer loyalty, Tarnowska and Ras (2023) \cite{Tarnowska2021} developed CLIRS2, an NLP-powered recommender system to extract actionable insights from unstructured text data. %By applying opinion mining and action rule mining algorithms, the system identified the minimal interventions required to maximize satisfaction and profitability. Their framework, tested in repair service comments, demonstrated superior capabilities in processing large-scale feedback compared to traditional methods.

Shahin et al. (2024) \cite{Shahin2024} showcased GPT-3.5 Turbo's effectiveness in extracting nuanced multilingual insights for Voice of Customer (VoC) analysis, integrating NLP with Lean Six Sigma 4.0 to support real-time service strategies. Shu et al. (2024) \cite{Shu2022} applied NLP to analyze social media feedback on electric vehicles, identifying key perceived risks—such as Performance and Time Risk—with over 40\% negative sentiment in all categories. Huang et al. (2022) \cite{Huang2022} highlighted the potential of NLP-powered chatbots in improving communication and engagement, particularly within customer service and language learning applications.

%%%%%%%%%%%%%%%%%%%%%%%%%%%%%%%%%%%%%%%%%%%%%%%%%%%%%%%%%%% ISR

Recent advances in machine learning (ML) and natural language processing (NLP) have transformed customer feedback analysis, enabling organizations to extract actionable insights for personalized service delivery. Yang et al. (2023)\cite{Yang2023} demonstrated how Siamese TextCNN and attention mechanisms help to analyze customer sentiments to increase engagement. Bauer et al. (2023) \cite{Bauer2023} stressed the importance of explainable AI, showing that feature-based explanations improve user understanding but also risk biases and decision-making inconsistencies.% Together, these advancements highlight the growing intersection of data science and service optimization.

Despite extensive research on customer feedback in commercial sectors, there remains a notable gap in understanding and modeling service feedback within public tax administrations. This is especially relevant for organizations like the Canada Revenue Agency, which operate in complex, multilingual, and policy-driven environments. The existing literature rarely addresses the specific nature of taxpayer feedback or the administrative challenges faced by public sector institutions, making this a critical and underexplored area. Our research directly responds to this gap by developing a feedback analysis system tailored to the operational context of the Tax Service and of tax services in general.

In addition, several critical gaps remain, underscoring the need for our proposed solution. One of the primary limitations of existing research is the insufficient focus on multilingual feedback analysis. Although numerous studies have used NLP and machine learning techniques, they often do not address the complexities involved in processing multilingual feedback. Current approaches struggle with the simultaneous analysis of feedback in multiple languages, limiting their applicability in diverse linguistic environments.
Another limitation in current research is the absence of automated trend detection and analysis within feedback systems. Although there are various methods for topic and sentiment analysis, they rarely incorporate automated mechanisms to identify emerging, persistent, and disappearing trends. An end-to-end system capable of categorizing feedback while simultaneously tracking trends over time is still lacking in the literature.
Lastly, the integration of domain experts' experience with LLM-generated output remains underexplored, particularly in multilingual feedback analysis within public sector applications. Although some studies emphasize the value of combining expert judgment with machine learning models %\cite{liu2023}, 
this approach has not been extensively applied in public service feedback analysis. Our research addresses this gap by evaluating topic categorization using two complementary methods: a similarity score and a statistical comparison of human versus model-assigned scores. This dual validation approach ensures a more rigorous assessment of system performance and enhances the reliability of AI-assisted feedback analysis.
%By addressing these critical research gaps, our work aims to advance the field of feedback analysis, particularly in multilingual and public sector contexts, by integrating fine-tuned AI models with collaborator expertise to create a more accurate, efficient, and interpretable system.

\section{Methodology}

%A key component of our methodology is the incorporation of human-in-the-loop processes at multiple stages of development. Human oversight was first implemented during the de-identification stage, where privacy officers ensured that all sensitive information was appropriately masked while maintaining contextually relevant content necessary for downstream modeling. In addition, expert reviewers played a critical role during model validation, assessing the accuracy and relevance of the topic assignments generated by the LLMs. This dual stage involvement, during both preprocessing and evaluation, ensures the trustworthiness, fairness, and accountability of the system and aligns with ethical AI principles.

Our methodology includes human oversight during both data de-identification and model validation, ensuring fairness, trust, and accountability in line with ethical AI standards.

\subsection{Data Collection \& Preprocessing}

The dataset for this project was sourced from %multiple decentralized systems within the Tax Service,
 multiple decentralized systems within the Tax Service, all of which were approved for use from both privacy and legal standpoints. Although this presented substantial challenges in terms of data integration and consistency, rigorous care was taken to uphold data protection standards. Beyond these compliance considerations, it is essential to emphasize that organizations like the Tax Service must also systematically understand service feedback to drive continuous quality improvement and enhance the responsiveness of public service delivery. The collected dataset comprises both structured numerical features, such as categorical data and demographic attributes, and unstructured free-form text entries, mainly consisting of taxpayer feedback. This diverse mix of data types required customized preprocessing strategies to maximize the utility of each feature for analysis.

\subsubsection{Text Data}

\begin{figure}[!ht]
\centering
\includegraphics[width=3.5in]{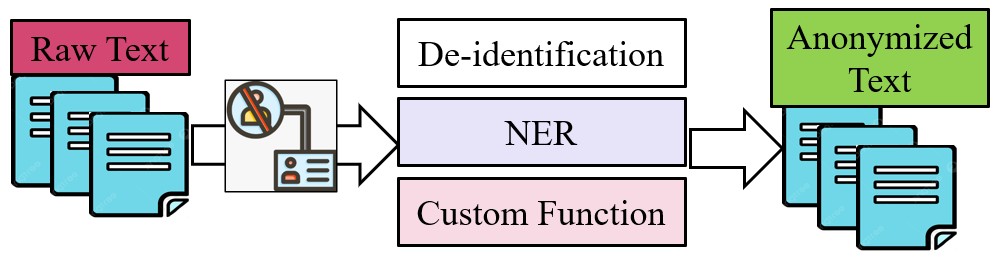}
\caption{Overview of Text Preprocessing Pipeline: De-identification, Named Entity Recognition (NER), and Custom Functions for Ensuring Privacy }
\label{fig:pre}
\end{figure}

As illustrated in Figure~\ref{fig:pre}, preprocessing steps were necessary to prepare the raw taxpayer feedback for analysis. A crucial component of this process was to ensure data security and privacy compliance, particularly in the handling of sensitive information.
Due to the sensitive nature of taxpayer feedback, which frequently contains personally identifiable information (PII), the first and foremost step was a thorough de-identification procedure. This was essential not only to comply with stringent data protection regulations, but also to facilitate secure data transfer to computational resources with greater processing power (e.g., enhanced GPU capabilities) for further analysis.
The de-identification process was comprehensive, involving several key actions: detecting and masking personal information such as Social Insurance Numbers and phone numbers; anonymizing individual names to prevent identity disclosure; masking or removing references to specific organizations and locations while retaining generalized geographic data (e.g., provinces); altering structured data like agent numbers and specific monetary amounts to prevent reidentification risks; and identifying and masking email addresses, web links, and other personal identifiers. In addition, certain numerical data were masked, although information such as percentages was retained to ensure the integrity of the statistical analysis. This rigorous process balanced the need for privacy with the preservation of data's analytical value. This process was also conducted to ensure that the model processing the feedback is less prone to bias, as no mention of a taxpayer's PII will be in the feedback that is fed to the LLM.
%%%
%Although the dataset was not explicitly labeled biased or unbiased, our pre-processing steps were designed to mitigate potential sources of bias and support fairness in the downstream modeling. In particular, the removal of personally identifiable information (PII) helps prevent the model from learning unintended associations between personal identifiers (e.g., names, cities, or agent IDs) and specific feedback topics. This reduces the risk of overfitting to identity-linked patterns that could skew topic classification. While we did not formally capture evaluation metrics, we implemented a human-in-the-loop validation process. A randomly selected sample of feedback was jointly reviewed to confirm the effectiveness. Based on this manual review, the results were  acceptable by Tax Service experts for the purposes of protecting privacy and maintaining fairness.Although explicit bias measurement is not performed at the data collection stage, our methodology identifies potential disparities at the analysis stage by stratifying data across demographic variables such as gender, age, and language. By comparing how topic relevance evolves across these groups, we assess whether certain demographic segments experience disproportionate issues - an approach aligned with fairness and bias-aware analysis in multilingual public service feedback.
%%%

We employed two approaches and tools, specifically utilizing the following methods for de-identification purposes.

\textbf{Advanced Entity Recognition Using Transformers:} To de-identify sensitive personal information within taxpayer feedback, we employed a methodology based on Transformer models, leveraging the powerful capabilities of BERT. Specifically, we utilized the \texttt{XLM-Roberta-Large-Finetuned-Conll03-English} model, a variant of RoBERTa (Robustly Optimized BERT Pretraining Approach) that has been fine-tuned over the CoNLL-03 dataset for Named Entity Recognition (NER). This model is particularly well suited for multilingual NER tasks, making it highly effective in processing both English and French texts, an essential capability given the bilingual nature of the feedback.
%According to published benchmarks on standard datasets such as OntoNotes 5.0, this model demonstrates high performance (e.g., precision ~0.97, recall ~0.95), enhancing the accuracy of identifying entities such as names, addresses, phone numbers and email addresses. The strong contextual understanding of the model allows it to adapt to variations in phrasing, synonyms, and complex sentence structures, improving the reliability of the anonymization process\cite{xlm-roberta-conll03}.%\href{https://huggingface.co/FacebookAI/xlm-roberta-large-finetuned-conll03-english}{XLM-RoBERTa Large Finetuned on CoNLL-03 English}.
However, recognizing that NER models alone may not capture all forms of personally identifiable information (PII), especially in noisy, informal, or domain-specific text, we complemented the transformer-based approach with additional rule-based pattern matching techniques. These included customized regular expressions to detect structured entities such as Social Insurance Number (SIN) formats, postal codes, and monetary values. Finally, we performed an expert review of the flagged examples to ensure a robust removal of sensitive information while preserving the contextual integrity of the feedback.

\textbf{Creating Functions for Specific Name Structures:}  In situations where our standard model did not fully obscure private information or when we required a more tailored approach to anonymization, we developed a custom function to enhance privacy protection based on the organization's specific needs. This function relies primarily on regular expressions to identify and anonymize patterns associated with financial figures, phone numbers, specific year formats, addresses, cities of employment, websites, email addresses, and certain personal names that follow identifiable patterns, along with various other personally identifiable information (PII).
Although the function is designed to mask sensitive details such as monetary amounts and years, it selectively retains nonsensitive data such as monetary percentages and month names to align with specific data retention requirements. Additionally, it effectively anonymizes names that may not be easily recognized by conventional models, as well as city names, while allowing country names to remain visible. This customized approach is usually necessary and industry-dependent. It ensures a more precise and adaptable de-identification process, meeting stringent data privacy requirements while preserving essential contextual information.

\subsubsection{Numeric Data}

\textbf{Service Quality Elements:} Each feedback entry is assigned at least one Service Quality Element topic, SQE, categorizing the feedback into specific service quality dimensions. This categorization has been performed manually. % tax experts
Tax Service Feedback Program Officers. The various SQE topics are illustrated in Figure~\ref{fig:sqe}.

\definecolor{mycolor}{RGB}{0,102,153} % Define your custom color

\begin{figure}[ht]
\centering
\begin{tikzpicture}
    % Row 1
    \node[draw, fill=mycolor, text=white, minimum width=2.5cm, minimum height=1.1cm, align=center] (A1) at (0,0) {Access to \\ escalation};
    \node[draw, fill=mycolor, text=white, minimum width=2.5cm, minimum height=1.1cm, align=center] (A2) at (3,0) {Accessibility };
    \node[draw, fill=mycolor, text=white, minimum width=2.5cm, minimum height=1.1cm, align=center] (A3) at (6,0) {Availability};
    \node[draw, fill=mycolor, text=white, minimum width=2.5cm, minimum height=1.1cm, align=center] (A4) at (9,0) {Clarity};
    \node[draw, fill=mycolor, text=white, minimum width=2.5cm, minimum height=1.1cm, align=center] (A5) at (12,0) {Completeness};
    
    % Row 2
    \node[draw, fill=mycolor, text=white, minimum width=2.5cm, minimum height=1.1cm, align=center] (B1) at (0,-1.3) {Consistency};
    \node[draw, fill=mycolor, text=white, minimum width=2.5cm, minimum height=1.1cm, align=center] (B2) at (3,-1.3) {Convenience};
    \node[draw, fill=mycolor, text=white, minimum width=2.5cm, minimum height=1.1cm, align=center] (B3) at (6,-1.3) {Findability};
    \node[draw, fill=mycolor, text=white, minimum width=2.5cm, minimum height=1.1cm, align=center] (B4) at (9,-1.3) {Information \\ Accuracy};
    \node[draw, fill=mycolor, text=white, minimum width=2.5cm, minimum height=1.1cm, align=center] (B5) at (12,-1.3) {Information \\ Format};
    
    % Row 3
    \node[draw, fill=mycolor, text=white, minimum width=2.5cm, minimum height=1.1cm, align=center] (C1) at (3,-2.6) {Official \\ languages};
    \node[draw, fill=mycolor, text=white, minimum width=2.5cm, minimum height=1.1cm, align=center] (C2) at (6,-2.6) {Professionalism};
    \node[draw, fill=mycolor, text=white, minimum width=2.5cm, minimum height=1.1cm, align=center] (C3) at (9,-2.6) {Timeliness};
\end{tikzpicture}

\caption{The 13 Service Quality Elements in which each feedback is classified into. As opposed to the current Tax Service process, each piece of feedback can be associated with more than one SQE.}
\label{fig:sqe}
\end{figure}
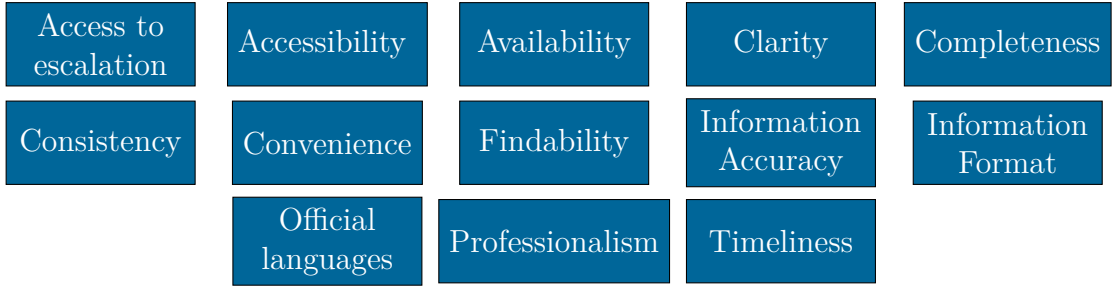

\textbf{Demographic Features:} For each feedback category, we also capture demographic information, including 'Gender', 'Preferred Language' and 'Age.' These are the demographic features for which we will track emerging topics within the feedback texts.

\subsection{General framework}

\begin{figure}[!h]
\centering
\includegraphics[width=0.8\textwidth]{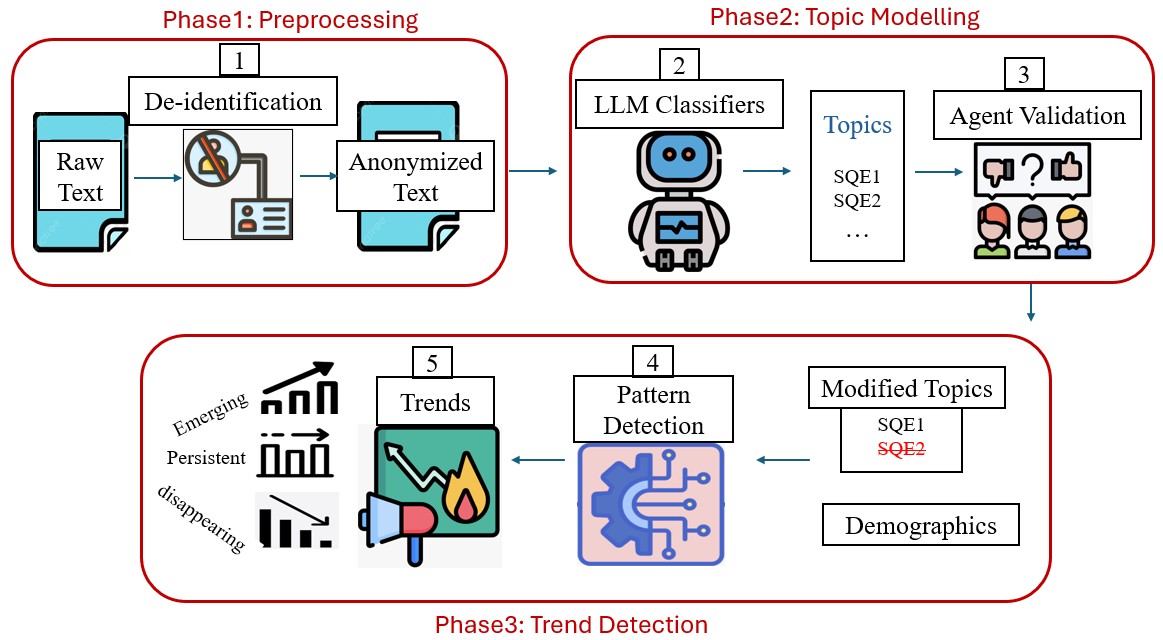}
\caption{Flowchart showing the process of analyzing de-identified text feedback to detect trends across demographics. Steps include de-identification, LLM classification, expert validation, pattern detection, and trend analysis. This process adheres to Ethically aligned AI principles, ensuring fairness, accountability, transparency, and explainability.}
\label{fig:flowchart}
\end{figure}

This study introduces a tool to analyze taxpayer feedback by classifying it into Service Quality Elements (SQEs) and detecting topic trends across demographic groups. Figure~\ref{fig:flowchart} outlines the overall methodology, while Sections~\ref{sec:Topic Modeling} and~\ref{sec:Trend Detection} detail technical components such as model tuning and trend analysis.

\begin{enumerate}
    \item \textbf{Preprocessing:}  
    Raw feedback is de-identified to protect privacy and eliminate bias linked to personal identifiers, supporting ethical AI practices.

    \item \textbf{Text Processing via LLM Classifiers:}  
    A fine-tuned and quantized LLM classifies feedback into SQEs with high accuracy and efficiency, capturing nuanced content in both English and French.

    \item \textbf{Agent Validation:}  
    Tax Service representatives validate model outputs, enhancing accountability and transparency, and helping refine the model for greater alignment with human judgment.

    \item \textbf{Trend Detection \& Pattern Analysis:}  
    Regression models identify whether SQE-related topics are emerging, persistent, or fading across demographics, revealing potential disparities and improving explainability.

    \item \textbf{Ethical Standard:}  
    The system emphasizes fairness, transparency, and human oversight. It detects unequal topic trends across groups using demographic stratification and regression. Explainability is ensured through justifications for each SQE assignment, enabling trust and accountability in public service applications.
   To reinforce accountability, expert oversight remains integral to the process, with experts retaining final responsibility for decisions—ensuring that a person, not just an algorithm, is accountable for the model’s outcomes. Transparency is equally critical; users are clearly informed about the AI’s role in processing and analyzing feedback, enabling informed engagement with system recommendations.
    Explainability, as defined in the ethical AI literature, refers to the ability of external users—such as stakeholders, or auditors—to understand and trust the model’s output. This contrasts with interpretability, which typically emphasizes internal teams understanding the model’s architecture or parameters \cite{Guidotti2019}. In our system, explainability is enhanced by generating explicit justifications for each topic and relevance score, allowing stakeholders to trace how feedback is categorized. 
    We align our framework with responsible AI design principles that emphasize fairness, transparency, and human agency, particularly in public sector applications \cite{Morley2021}. 
\end{enumerate}

\subsection{Topic Modeling}\label{sec:Topic Modeling}

The methodology aims to categorize taxpayer feedback into 13 predefined elements of service quality (Figure~\ref{fig:sqe}).% Given the large volume of feedback and thematic similarities among responses, initial attempts to use traditional and well-known topic modeling techniques, such as BERT \cite{muralitharan2024privacy} and LDA \cite{wahid2022topic2labels}, were not entirely effective. These methods often produced topics that overlapped across different texts, limiting their ability to provide clear and distinct categorizations. This challenge necessitated a more refined approach that could capture subtle nuances in feedback, closely resembling human analysis. Consequently, we explored the use of an advanced large language model. When utilizing LLMs, we considered two approaches: prompt engineering with a pre-trained model, which offers a cost-effective solution, and fine-tuning a more sophisticated model that can be customized to specific themes and organizational requirements. Based on the organization's model licensing constraints, resource availability, and model performance criteria, we selected two models for evaluation. Note that the model selection was done at the time of model deployment. Since then, other models have been released that improve performance over our choices. However, structurally our choices are generalizable: They are all transformer-decoder-based models, and they are considered to be small within the LLM range (in the 5B - 13B parameter range).
To classify feedback into 13 categories, we moved beyond traditional models like BERT and LDA because they often produced overlapping and indistinct topics, making it difficult to clearly assign feedback to specific service elements. Instead, we adopted a refined LLM-based approach, evaluating both prompt engineering and fine-tuning methods on small transformer-decoder models (5B--13B), selected based on organizational constraints and deployment timing.

\subsubsection{Prompt Engineering}
For prompt engineering, we use Zephyr-7B-$\beta$ \cite{zephyr7b}. Zephyr is a 7 billion parameter language model, similar in architecture to GPT models, and fine-tuned for natural language processing tasks. It is based on Mistral-7B-v0.1\cite{mistral7b}, an efficient and powerful Transformer-based model known for its strong language understanding capabilities.
Zephyr has been specifically fine-tuned on a diverse mix of publicly available and synthetic datasets, enabling it to generate high-quality text while maintaining broad generalizability across various applications. 
%%%
%In our use case, the taxpayer feedback consisted of English and French text.  Because we employ prompt engineering, the model is directly exposed to bilingual inputs during inference. This allows us to assess its responses in both English and French. In practice, the model demonstrates sufficient understanding of both languages to support multilingual feedback analysis.%%%
One of Zephyr's key advantages is its MIT license, which provides flexibility for both research and commercial use. This makes it an attractive option for developers and organizations seeking to integrate an advanced NLP model into their workflows without restrictive licensing constraints.

\subsubsection{Fine-Tuned LLM }
Mistral-7B-Instruct-v0.2 \cite{mistral7b-instruct} is an updated version with a greatly expanded context window size, from 8,000 to 32,000 tokens, which enhances its ability to handle longer text sequences. This model has moved away from the sliding-window attention mechanism to a more efficient design, which reduces computational costs and increases processing speed. Importantly, the Mistral model has been trained multilingually, making it capable of understanding and processing inputs in both English and French, an essential feature for analyzing Canadian taxpayer feedback.
We chose to fine-tune this model to compare its performance against the prompt engineering approach. Our goal is to adapt the model to better handle the task using its enhanced architecture, which includes a larger context window and optimized attention mechanisms. These improvements make it particularly well suited for analyzing complex and lengthy texts, ensuring more accurate and context-aware processing.
By refining it with authority-specific data and expert input, we improve its ability to detect subtle patterns and specific nuances in taxpayer feedback, which generic models might overlook. This customization not only ensures a more in-depth understanding of the diverse feedback but also aligns the model more closely with operational requirements and standards, thus improving the accuracy and efficiency of our service quality evaluations.

\subsubsection{Quantize fine-tuned model using GPTQ:} 
After fine-tuning the Mistral-7B-Instruct-v0.2 we decided to quantize it using Gradient-Preserving Quantization (GPTQ) \cite{Frantar2023}.
Gradient-Preserving Quantization (GPTQ) is a technique designed to reduce the computational complexity of large neural networks without significantly affecting their performance. By quantizing the weights of the model to lower precision, such as reducing 32-bit floating point numbers to 16-bit or 8-bit, GPTQ helps decrease the model's memory footprint and speeds up processing. Unlike traditional quantization methods that can lead to a loss in precision, GPTQ maintains the effectiveness of the model by carefully adjusting weight reductions in a way that minimizes the impact on gradient flows during training \cite{frantar2022gptq, smith2022advancedquant}. This makes GPTQ particularly useful for deploying complex models on devices with limited computational resources, maintaining a balance between efficiency and performance.

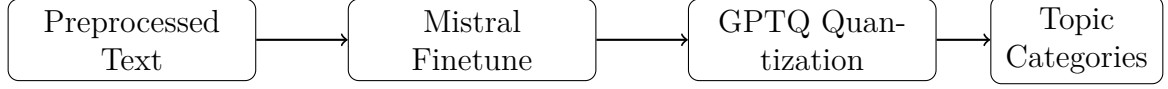
\begin{figure}[h]
\centering
\begin{tikzpicture}[node distance=0.5cm, auto]
  % Nodes
  \node (Tax Service) [rectangle, draw, fill=white!20, text width=3cm, text centered, rounded corners, minimum height=1cm] {Preprocessed Text};
  
  \node (mistral) [rectangle, draw, fill=white!20, text width=3cm, text centered, rounded corners, minimum height=1cm, right of=Tax Service, xshift=4cm] {Mistral Finetune};
  \node (gptq) [rectangle, draw, fill=white!20, text width=3cm, text centered, rounded corners, minimum height=1cm, right of=mistral, xshift=4cm] {GPTQ Quantization};
  \node (tmmodel) [rectangle, draw, fill=white!20, text width=2cm, text centered, rounded corners, minimum height=1cm, right of=gptq, xshift=3cm] {Topic Categories};

  % Arrows
  \draw[->, thick] (Tax Service) -- (mistral);
  \draw[->, thick] (mistral) -- (gptq);
  \draw[->, thick] (gptq) -- (tmmodel);
\end{tikzpicture}
\caption{Flowchart illustrating the processing of the preprocessed text through the model, which is then fine-tuned and subsequently quantized using GPTQ before being utilized for Topic Modeling purposes.}
\label{fig:flowchart1}
\end{figure}

This approach was chosen to enhance the model's resource efficiency without significantly affecting its performance. Quantization minimizes the computational load of the model by reducing the precision of the data it processes, which decreases memory requirements and speeds up response times. Implementing GPTQ ensures that these resource reductions do not undermine the model's ability to recognize subtle feedback differences, which is critical for the detailed analysis necessary in our work. This strategy enables Mistral-7B-Instruct-v0.2 to operate more efficiently on the organization's existing infrastructure, facilitating a more sustainable and cost-effective implementation while maintaining robust performance.
The overall workflow, from preprocessing to topic modeling, is depicted in Figure~\ref{fig:flowchart1}.

\subsection{Trend Detection Analysis}
\label{sec:Trend Detection}
After extracting topics and digitizing relevance scores, the methodology progresses to its final stage: analyzing emerging trends using logistic regression. This approach allows for a more in-depth understanding of how specific topics have evolved across feedback received while considering key demographic variables. The analysis focuses on categorical factors that were selected as the most critical to the organization's operation: age group, gender, and preferred language to uncover meaningful patterns within the feedback.
Through this method, topics are classified into three different trend categories based on their prominence and evolution. \textbf{Emerging trends} represent new topics that have recently surfaced in the feedback data, indicating evolving concerns. \textbf{Persistent trends} highlight topics that remain consistently significant in both time periods, demonstrating their ongoing relevance. In contrast, \textbf{disappearing trends} reflect topics that were initially prominent but have shown a decrease in frequency, suggesting a reduced level of concern or interest.

\begin{figure}[!ht]
\centering
\includegraphics[width=0.8\textwidth]{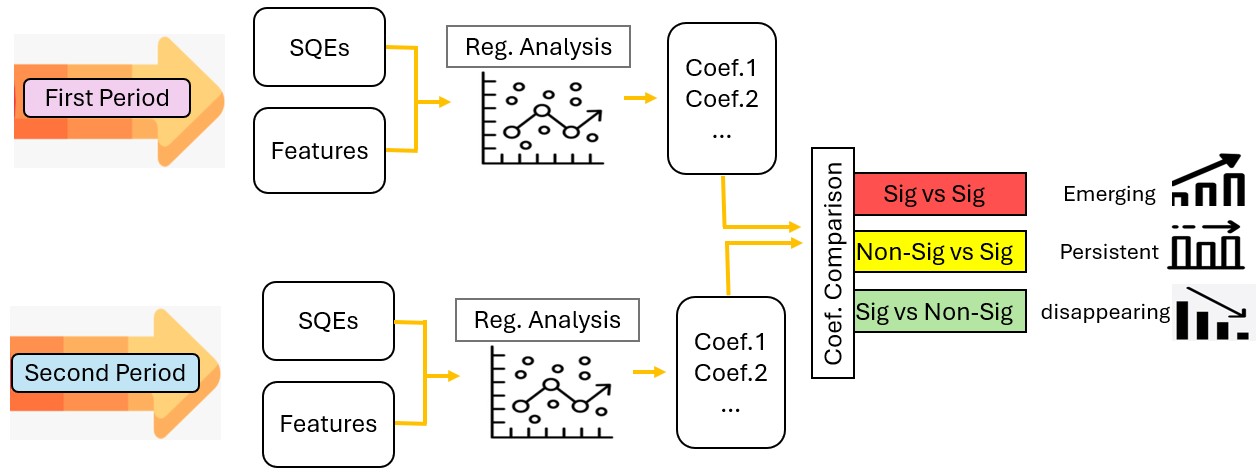}
\caption{Workflow of trend analysis across two time periods. The figure depicts the process of data splitting, feature extraction, logistic regression modeling, coefficient comparison, and the classification of trends into emerging, persistent, or disappearing categories.}

\label{fig:trend}
\end{figure}

\subsubsection{Timeframes for the Trend Analysis Process}
Our trend analysis begins with establishing a split date to divide the dataset into two distinct segments. Given that the feedback spans an entire year, we employ two different splitting strategies to examine the evolving dynamics of taxpayer concerns. These approaches help uncover significant shifts in feedback trends while balancing sensitivity to short-term changes and broader patterns over time.

\paragraph{Entire Year Versus New Quarter}  
An approach involves using the entire year as a baseline period and comparing it with the most recent quarter. This method allows us to detect notable changes in feedback themes, revealing whether new concerns have emerged or if existing ones have gained or decreased in prominence versus the operation during the previous year. By analyzing a full year of data as context, we ensure that trends are assessed comprehensively. 
However, this approach may obscure subtler or emerging signals, as comparing a short recent period against a long baseline can dilute the visibility of weaker patterns that are only beginning to manifest.
%However, this approach comes with the challenge of higher variability, as each quarter can exhibit significant differences compared to the previous year, which can lead to fluctuations in detected trends.

\paragraph{Semester Versus Last Semester}  
Another approach divides the year into two equal periods of six months. This split facilitates a direct comparison between the first and second halves of the year, enabling a more precise analysis of changes in feedback patterns. However, this method may introduce additional noise, as tax declarations and related interactions may vary across different times of the year due to seasonal effects. An alternative strategy, if more data were available, would be to compare equivalent semesters over consecutive years. This would help determine whether issues from the last semester persist into the following year, offering a more reliable way to assess long-term trends while mitigating seasonal distortions.

By applying these trend analysis techniques, we gain a clearer understanding of how feedback evolves, ensuring that emerging patterns are detected with both short-term responsiveness and long-term stability.

\subsubsection{Logistic Regression Model}
The deidentified and privacy-compliant dataset is then prepared for logistic regression analysis. 
%%%
%It is noted that we selected logistic regression over alternatives such as time series models because the objective of this stage was not to forecast topic frequencies over time, but rather to assess statistical differences in topic relevance across demographic groups between two defined time periods. Logistic regression provides a transparent and interpretable way to measure associations between topic occurrence and categorical demographic variables, allowing us to calculate confidence intervals and determine statistical significance. Moreover, logistic regression is computationally efficient and well-suited for our stratified analysis approach, which emphasizes fairness and explainability in trend detection. The modeling involves one-hot encoding of categorical features such as 'Gender', 'Age group', and 'Language', transforming them into a format suitable for the logistic regression model. Separate encodings are applied for the early and late periods to maintain the temporal distinction.
%The objective variable in each model is each of the topics, and the independent variables are the categories that are identified as significant to monitor (Age, Gender, Language). This multiclass approach allows us to model the relationship between the encoded features and the identified topics, generating coefficients that represent the influence of each feature on the likelihood of each topic. We specifically use the standard coefficients of the regression, as they will be independent of the specific topic composition within the period under study. %%%
%%%

Two sets of logistic regression models are trained separately on feedback from earlier and later periods to assess how demographic influences on topic relevance change over time. For each SQE topic, the dependent variable \( Y_i \in \{0, 1\} \) indicates whether the topic was present in feedback \(i\). Demographic features \( X_{ij} \) serve as predictors in estimating the coefficients \( \beta_j \), representing their effect on topic occurrence:

\begin{equation}
\log\left( \frac{P(Y_i = 1)}{1 - P(Y_i = 1)} \right) = \beta_0 + \sum_{j=1}^{k} \beta_j X_{ij}
\end{equation}

To detect evolving patterns, coefficients from the two time periods are compared:

\[
\Delta \beta_j = \beta_j^{(2)} - \beta_j^{(1)}
\]

This difference captures the change in influence of each demographic feature over time, helping identify emerging, persistent, or fading topic trends.

\subsubsection{Trend Detection}

To assess changes in demographic influence on topic trends, we apply a bootstrapping method by resampling the dataset thousands of times and recalculating topic coefficients. This process yields distributions of coefficient estimates, from which we compute mean values and 95\% confidence intervals. 
A topic is considered \textit{emerging} if a feature becomes statistically significant in the current period, \textit{disappearing} if it loses significance, and \textit{persistent} if it remains significant across both periods. Features that are insignificant in both timeframes are excluded. This method enables a robust, statistically grounded analysis of evolving feedback patterns.

\section{Result}
\subsection{Text visualization}

\begin{figure}[ht]
    \centering

    % First row: ngrams_en1 and ngrams_en2 side by side
    \subfloat[Plot 1: Top 10 Most Frequent Unigrams in English Feedback\label{fig:ngrams_en1}]{
        \includegraphics[width=0.45\textwidth]{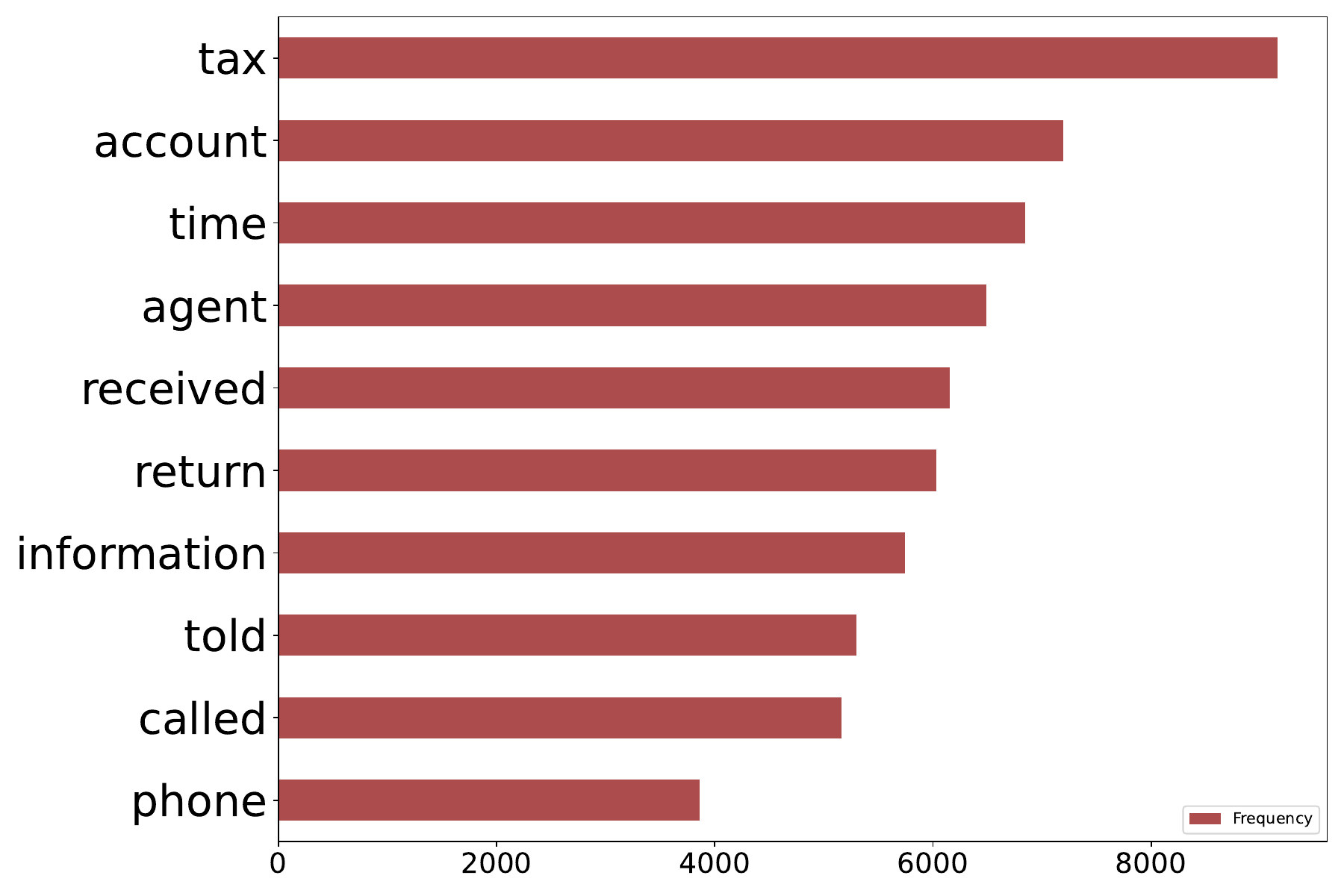} % Replace 'ngrams_en1.pdf' with the actual path
    }
    \hfill % Horizontal space between subfigures
    \subfloat[Plot 2: Top 10 Most Frequent Bigrams in English Feedback\label{fig:ngrams_en2}]{
        \includegraphics[width=0.45\textwidth]{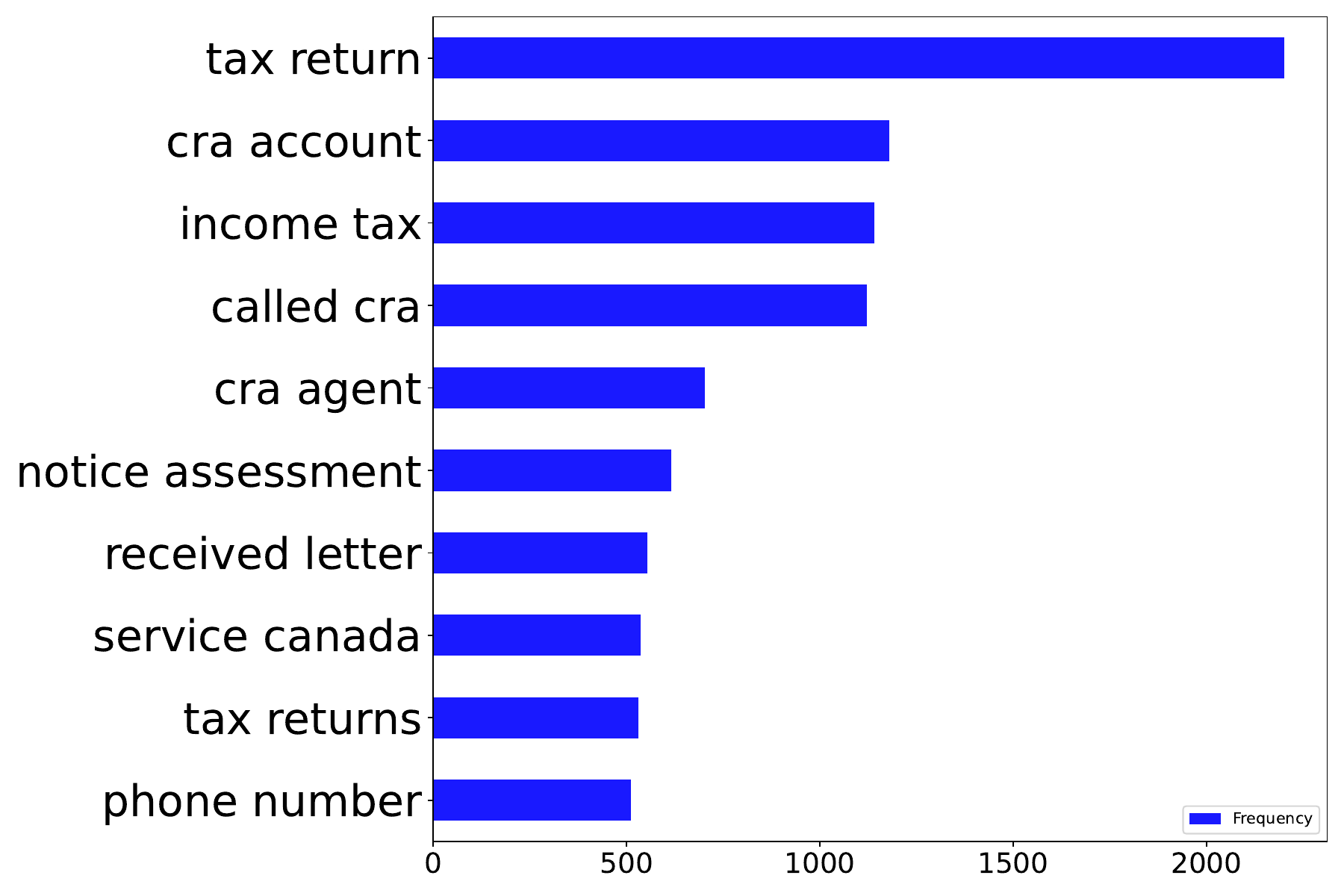} % Replace 'ngrams_en2.pdf' with the actual path
    }
    
    %\vspace{1cm} % Vertical space between the top and bottom rows
    
    % Second row: ngrams_fr1 and ngrams_fr2 side by side
    \subfloat[Plot 3: Top 10 Most Frequent Unigrams in French Feedback\label{fig:ngrams_fr1}]{
        \includegraphics[width=0.45\textwidth]{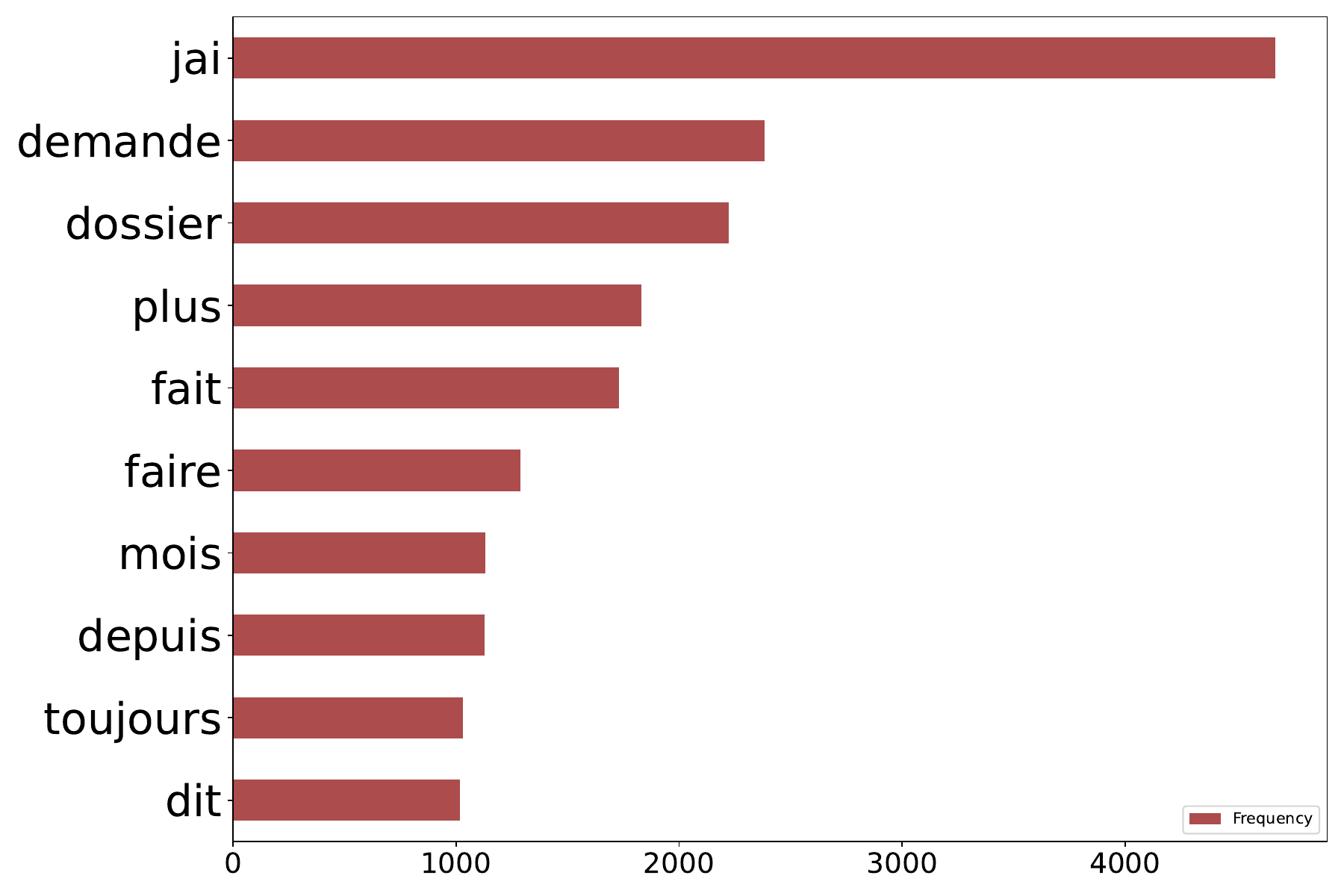} % Replace 'ngrams_fr1.pdf' with the actual path
    }
    \hfill % Horizontal space between subfigures
    \subfloat[Plot 4: Top 10 Most Frequent Bigrams in French Feedback\label{fig:ngrams_fr2}]{
        \includegraphics[width=0.45\textwidth]{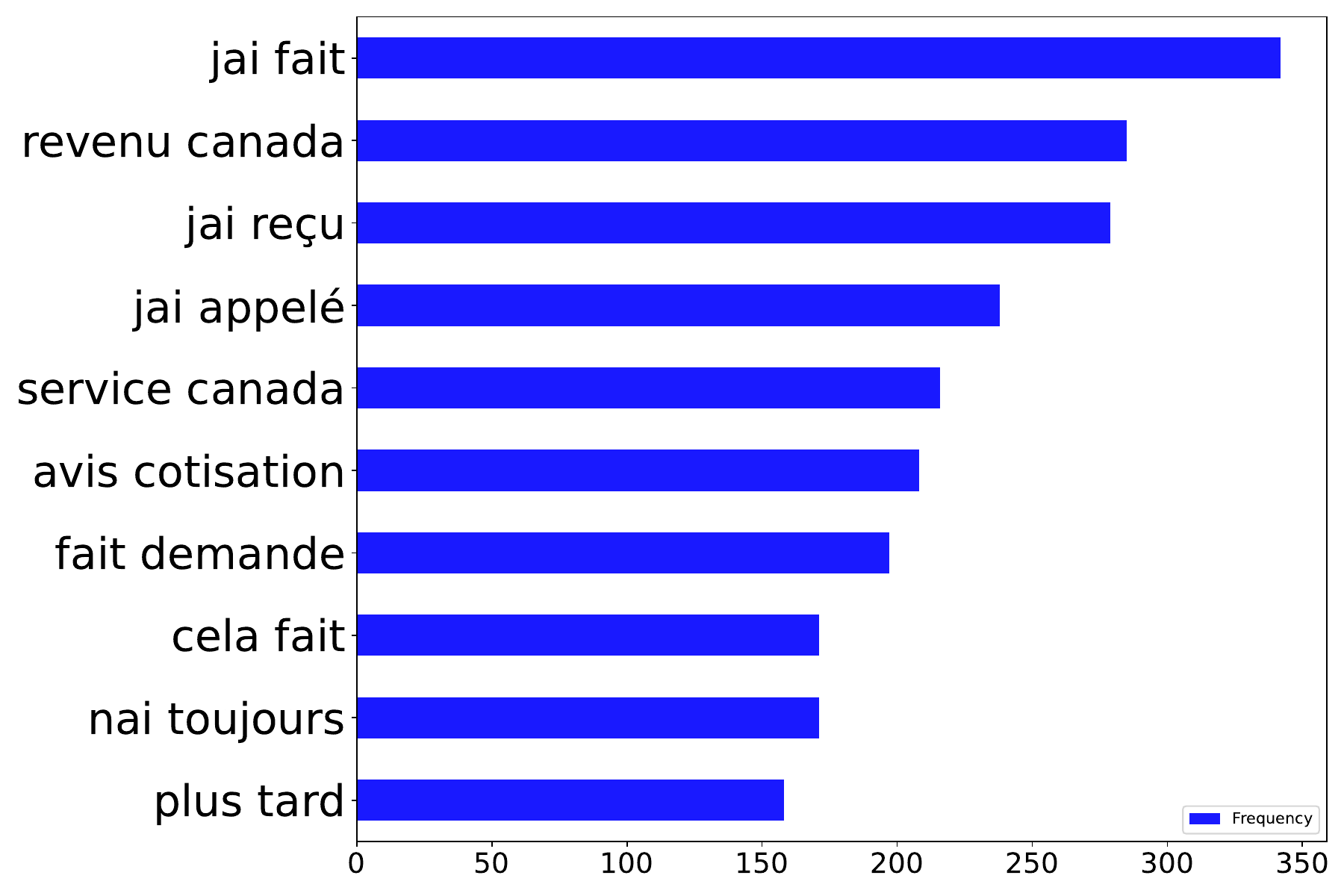} % Replace 'ngrams_fr2.pdf' with the actual path
    }

    \caption{Unigram and Bigram Frequency Analysis of English and French Feedback}
    \label{fig:ngrams_all}
\end{figure}

We have collected a dataset consisting of 6,515 feedback records in English and 1,646 records in French. To gain a better understanding of the patterns, themes, and key topics present in both languages, we employed a unigram and bigram\cite{garg2022ubis} approach for text analysis. Unigrams refer to individual words in the feedback, and analyzing them helps us capture the most frequent words or terms that occur across all records. Bigrams, on the other hand, consider pairs of consecutive words, allowing us to capture common phrases or word associations that convey more context or meaning than single words. 
In Figure~\ref{fig:ngrams_all}, we observe four graphs (labeled a, b, c, and d) that illustrate the unigram (graphs a and c) and bigram (plots b and d) analyses from the English and French feedback datasets.
Plots (a) and (c) show unigram frequencies with the most common words that appear on the text. For instance, terms like `tax' and `account' are dominant in the English feedback, while words like `\emph{jai}', `\emph{demande}' and `\emph{dossier}' are prominent in the French feedback. 
Plots (b) and (d) show the most frequent bigrams, revealing common-word pairs. In English, phrases such as ``tax return'' and ``Tax Service account'' reflect key user concerns, whereas in French, expressions like ``jai fait'' suggest frequent references to receiving documents or services.
%%%This indicates that users in both languages often discuss issues related to tax returns, accounts, and service requests. But there may be a variation of topics in both languages.
%While it is not surprising that both language groups raise core tax-related topics such as returns and account access, differences in term usage and emphasis may reflect distinct patterns of interaction or service needs.%%%
This variation between the English and French datasets may suggest a potential bias in the concerns and feedback provided by English-speaking and French-speaking taxpayers. This raises the need for further investigation in subsequent sections to explore whether differences in feedback reflect differing experiences or expectations.

\subsection{Topic Modeling Result}

As discussed in the methodology section, one step in the research process involves assigning topics (Figure~\ref{fig:sqe}) to the text. For this, we applied two approaches: prompt engineering using the Zephyr-7B-beta model and a quantized, fine-tuned LLM based on Mistral-7B-Instruct-v0.2. The following sections detail the implementation and results of these methods.

To effectively leverage a pre-trained model, we designed specific prompts in both English and French to accommodate the bilingual nature of the feedback. We used the Zephyr-7B-beta model, instructing it to categorize feedback based on key topics (SQEs) using relevance scores derived from its responses.
For each feedback entry, we structured an instruction set that combined an initial ``system'' directive with the user-provided content (i.e., the actual feedback). This was formatted in a conversational manner to improve the model’s comprehension and response accuracy.
The model was configured with a token limit of 2048 and parameter settings designed to balance response diversity while maintaining relevance. %Table \ref{tab:zephyr} in Appendix C summarizes these settings. 
After generating responses, we post-processed the output to extract only the relevant sections, compiling a categorized collection of feedback for further analysis.

%code: colab:englishLLM

%code: EnglishDigitalization

\subsubsection{LLM: Fine-Tuned \& Quantized for Efficiency}

\begin{figure}[!ht]
\centering
\includegraphics[width=\textwidth]{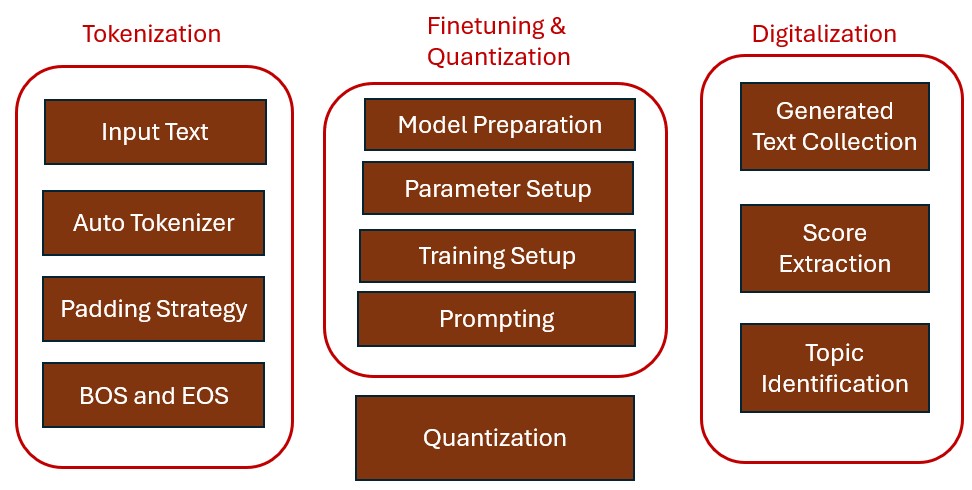}
\caption{Three-phase methodology for taxpayer feedback categorization: Tokenization prepares text input for processing; Fine-tuning adapts the model to Tax Service-specific SQEs; Digitalization organizes generated outputs into structured data for trend analysis and insights extraction.}
\label{fig:topic}
\end{figure}

To enhance the precision of topic modeling in taxpayer feedback, we fine-tuned the Mistral model specifically for the Tax Service's context. 
For this purpose, initially, a detailed prompt is fed into the model along with the feedback, outlining the SQEs and instructing the LLM to assign relevance scores and provide justification for each category. %The prompts themselves can be seen in Appendix A (English) and Appendix B (French). 
The LLM then processes this information to output a relevance score ranging from 1 to 5 for each SQE, accompanied by a word or sentence from the taxpayer feedback that justifies the score, thereby establishing a direct correlation between the feedback and specific service elements.
Subsequently, the output text containing the categorized feedback and relevance scores is parsed and structured into a data frame for further analysis. This process is repeated for each piece of feedback in the dataset, accumulating a comprehensive output that categorically assigns relevance scores to the feedback according to the predefined SQEs.
In extending our methodology to embrace Canada's bilingual landscape, the same approach is applied to French taxpayer feedback, utilizing a corresponding French prompt.% (Appendix B).
The flow of the process is visualized in Figure~\ref{fig:topic}, which divides the methodology into three primary phases: Tokenization, Fine-tuning, and Digitalization.

\subsubsection*{Phase 1: Tokenization}

This study employed the Mistral-7B-Instruct-v0.2 model, utilizing Hugging Face’s \texttt{AutoTokenizer} to convert textual input into tokens for model processing. Padding was applied to the right-hand side of sequences to standardize input lengths for batch processing. The end-of-sequence (EOS) token served as the padding token to prevent the model from learning from padded content. Additionally, special tokens—begin-of-sequence (BOS) and EOS—were added to mark sequence boundaries, aiding the model in identifying input structure during training.

%Tokenization is a preprocessing step in the natural language processing (NLP) pipeline, as it converts raw text into a format that can be understood by machine learning models. 
%For this study, the Mistral-7B-Instruct-v0.2 model was used, using Hugging Face's AutoTokenizer to handle the tokenization process. The tokenizer is an essential component, as it translates textual input into tokens that represent words, sub-words, or even characters. These tokens serve as the basis for the further processing of the input data so that the model can use them.
%An important configuration applied during tokenization is the padding strategy. Padding ensures that input sequences of varying lengths are standardized before being passed to the model, enabling batch processing. The padding in this case was configured to align with the right-hand side of the sequence. This means that when an input sequence is shorter than the required length, padding tokens are added to the right until the desired length is achieved. 
%In this configuration, the end-of-sequence (EOS) token was selected as the padding token. This token signals to the model that the meaningful part of the input has concluded, ensuring that the model does not inadvertently learn from the padded parts of the sequence. In addition to padding, special tokens such as the begin-of-sequence (BOS) token and the EOS token were introduced at the start and end of sequences, respectively. These tokens demarcate the boundaries of the sequences, helping the model to identify the start and end points during processing.

\subsubsection*{Phase 2: Fine-Tuning}

The fine-tuning of the Mistral model involved several key steps aimed at optimizing both performance and resource efficiency. To achieve this, the Low-Rank Adaptation (LoRA) technique was used in conjunction with parameter-efficient fine-tuning methods. These configurations were designed to strike a balance between maintaining model accuracy and reducing computational demands, particularly by applying k-bit training to the base model and selectively fine-tuning only certain parameters.

Parameter-Efficient Fine-Tuning (PEFT) was adopted to minimize computational complexity and memory usage. By fine-tuning only a subset of the model's layers, the LoRA method significantly reduces training overhead while preserving task-specific performance. This allows the model to efficiently adapt to new tasks without requiring full model retraining.
As illustrated in Figure~\ref{fig:topic}, the middle section focuses on fine-tuning. The fine-tuning process consists of several key phases. The Model Preparation phase involved configuring the base model for k-bit training, reducing both memory usage and computational overhead.
%%%In the LoRA Configuration phase, low-rank adaptation was applied to specific model layers, enabling fine-tuning of only essential parameters, thereby enhancing both memory efficiency and training speed. 
%In the LoRA Configuration phase, low-rank adaptation was applied to specific model layers, specifically the query and value projection layers within each Transformer block. These layers were chosen because they play a key role in the attention mechanism and are commonly targeted in parameter-efficient fine-tuning approaches. Focusing LoRA on these components enables the model to effectively adapt to new tasks with minimal training overhead, preserving performance while significantly reducing computational costs.%%%

For parameter setup, we used a numerical precision format known as nf4 (Normal Float 4), which represents a 4-bit floating point format optimized for neural network training. Unlike standard quantization formats, nf4 preserves more dynamic range and numerical stability, allowing efficient training while minimizing information loss\cite{dettmers2023qlora}. This format has been shown to perform well in low-precision training scenarios, especially when combined with techniques like LoRA. We also processed the data using a 16 bit brain floating point format (bfloat16) to process the activations and gradients during training. This 16-bit format strikes a balance between computational efficiency and numerical accuracy, especially in mixed-precision training setups. This hybrid use of nf4 and bfloat16 allows us to reduce memory consumption without significantly compromising model performance.
%A detailed summary of these configurations can be found in Appendix C (Table~\ref{param}).
During the Training Setup phase, the environment was optimized with critical hyperparameters, including the learning rate, batch size, and optimization strategy. Then, after the training setup, we designed a prompt to guide the model in generating structured responses aligned with Tax Service-specific SQEs.
%%%
%To ensure reproducibility and output stability during inference, we fixed the key generation parameters for the model. Specifically, the temperature was set to 0.0 and the top-p (nucleus sampling) to 1.0, eliminating randomness in the model output. This deterministic configuration ensured that, given the same input, the model consistently produced identical responses across multiple runs.%%%
%The configuration was selected based on initial experimentation and best practices from prior literature, prioritizing output stability and alignment with expert-annotated data. While we did not conduct a formal sensitivity analysis across decoding parameters, qualitative observations during development indicated that higher temperatures or sampling-based settings (e.g., top-k) led to greater variability and reduced alignment with expert-verified references.. 

\textbf{Quantization:}
To enhance the efficiency of our Large Language Model, we employed quantization, which reduces the model's memory footprint and accelerates inference without substantially compromising accuracy. Specifically, we used a 4-bit quantization configuration utilizing GPTQ (General Purpose Quantization) as the method. The bits parameter determines the bit width used for representing each weight in the neural network. %In our configuration, we set bits=4, which means that the model weights are quantized to 4-bit precision.
This drastically reduces the memory requirements and model size, as compared to the standard 16-bit (FP16) or 32-bit (FP32) floating point representations. A 4-bit quantization is a trade-off between accuracy and computational efficiency; while it introduces some approximation error, the reduction in memory usage allows the model to run faster and be loaded on devices with lower memory capacity, such as consumer GPUs.
\subsubsection*{Phase 3: Digitization}
After extracting topics and assigning relevance scores using our Large Language Model (LLM), the next step is to convert these scores into a structured format. This involves organizing the output into a dataframe for further analysis.
Since LLM generates text, we need to separate each piece of feedback into its key components: the assigned topic, the relevance score, and the reasoning behind the assignment.
At the core of this process is a function that uses regular expressions (regex) to identify patterns in the text.
%%%
%Regular expressions were used to detect structured segments in the model-generated text that match predefined Service Quality Elements. These patterns are designed to extract SQE labels and their corresponding confidence scores (ranging from 1 to 5), which the model explicitly returns in a standardized format (e.g., `SQE: Accessibility, Score: 0.82'). The relevance score is not computed externally but is a value generated by the model itself, reflecting its confidence in the SQE assignment. We consider an SQE relevant if its associated score exceeds a defined threshold, such as the mean or median of all possible scores across the dataset.%%%
This post-processing is applied to all feedback, resulting in a structured dataset with distinct columns for scores, topics, and justifications. 
%%%
%The Justification is directly extracted from the model-generated response using structured prompts. Each prompt instructed the model to provide not only the predicted SQE and its score but also a short rationale explaining why that SQE was relevant to the given feedback. This justification typically includes key phrases or themes from the feedback text that the model considered when assigning the SQE, offering transparency into the model's reasoning.%%%
This structured data enables further analysis, including trend detection.

\subsubsection{Model Evaluation}

\subsubsection*{ Topics Comparison}

\begin{figure}[!ht]
\centering
\includegraphics[width=\textwidth]{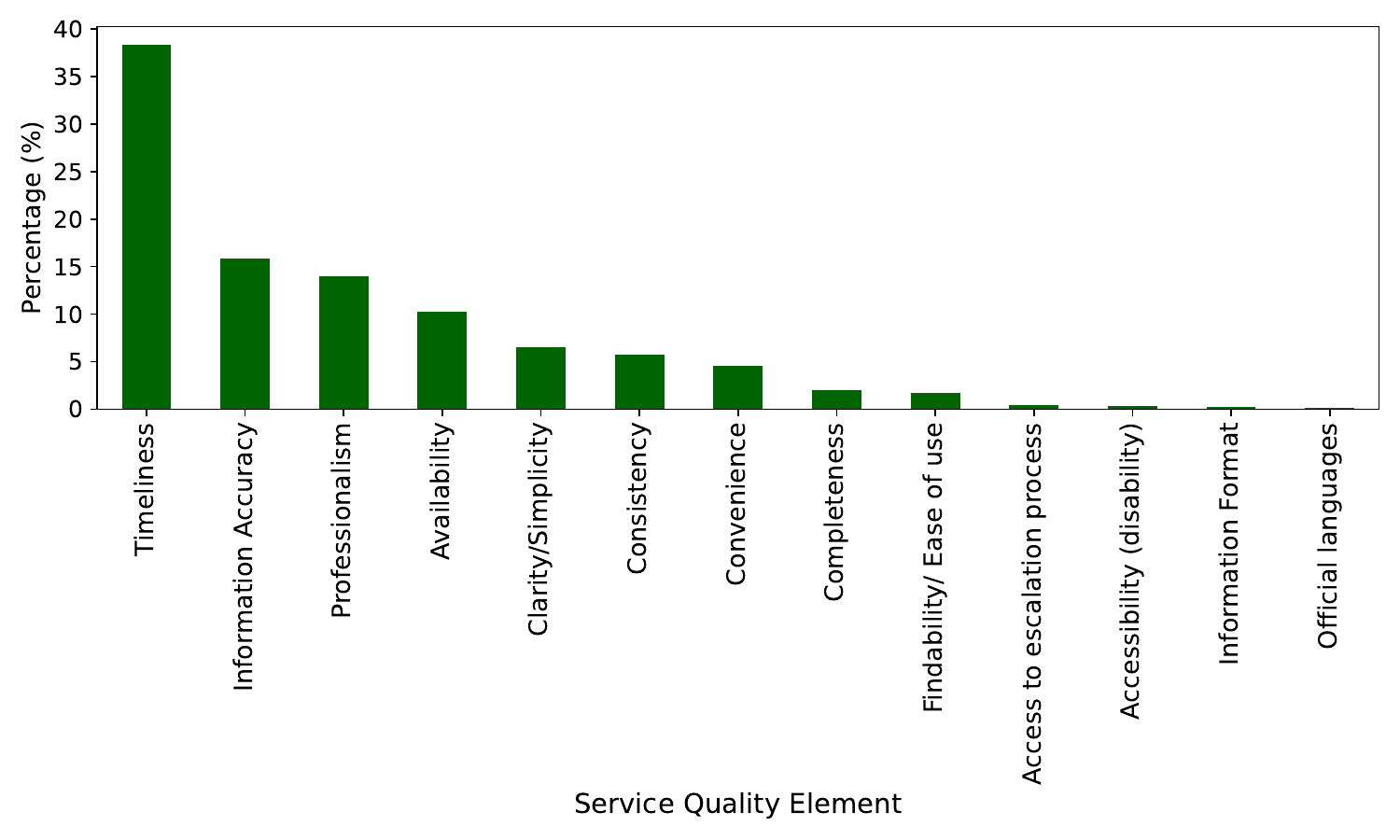}
\caption{Topic distribution across Service Quality Elements as categorized by Tax Service experts. The distribution shows a significant emphasis on  ``Timeliness'', indicating the prioritization of promptly actionable issues.}

\label{fig:Topic_distribution}
\end{figure}

\begin{figure}[!ht]
\centering
\includegraphics[width=\textwidth]{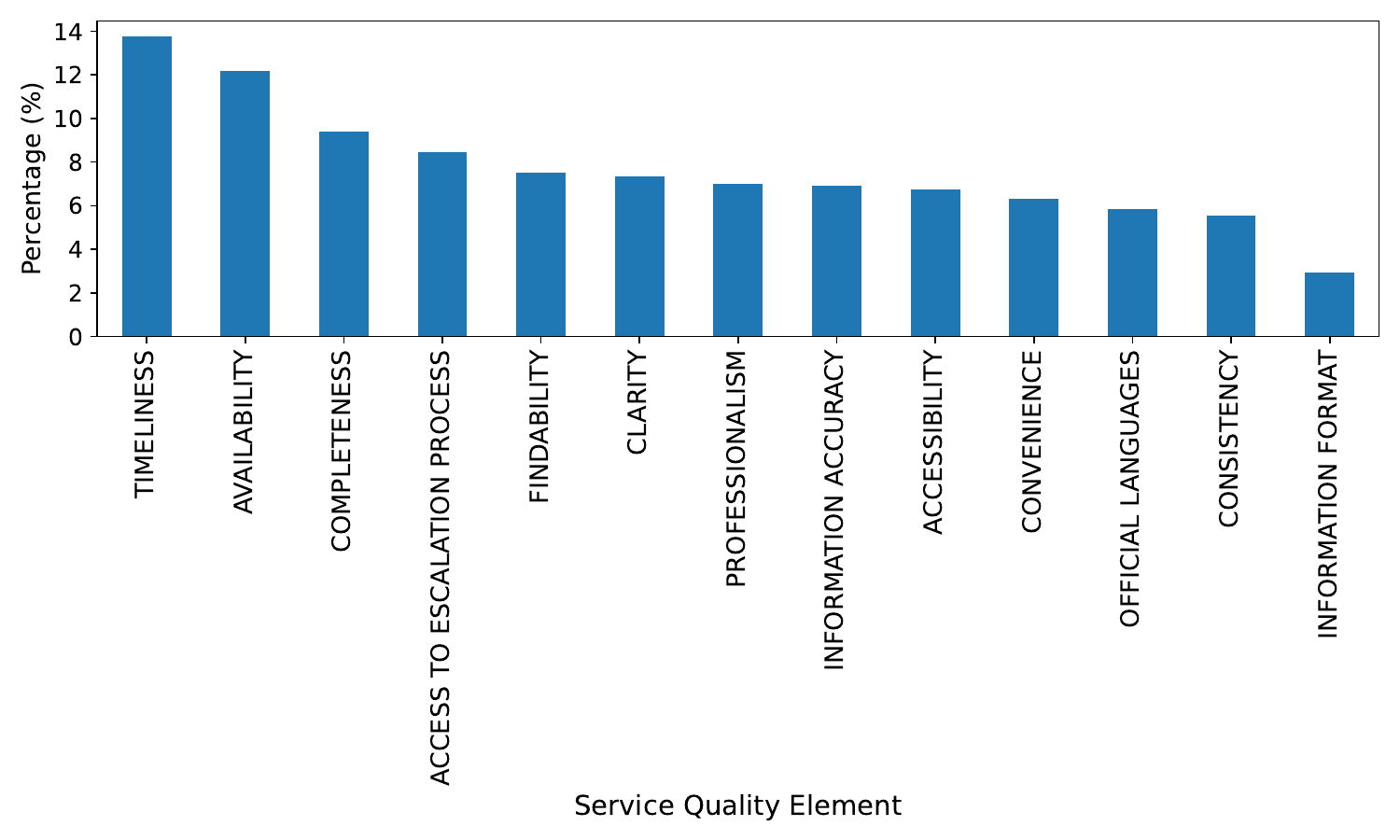}
\caption{Topic distribution across SQEs identified by the pretrained LLM (Zephyr). The results display a more balanced spread, highlighting the model's capability to recognize a wider range of issues in the feedback data.}

\label{fig:sqe_z}
\end{figure}

\begin{figure}[!h]
\centering
\includegraphics[width=\textwidth]{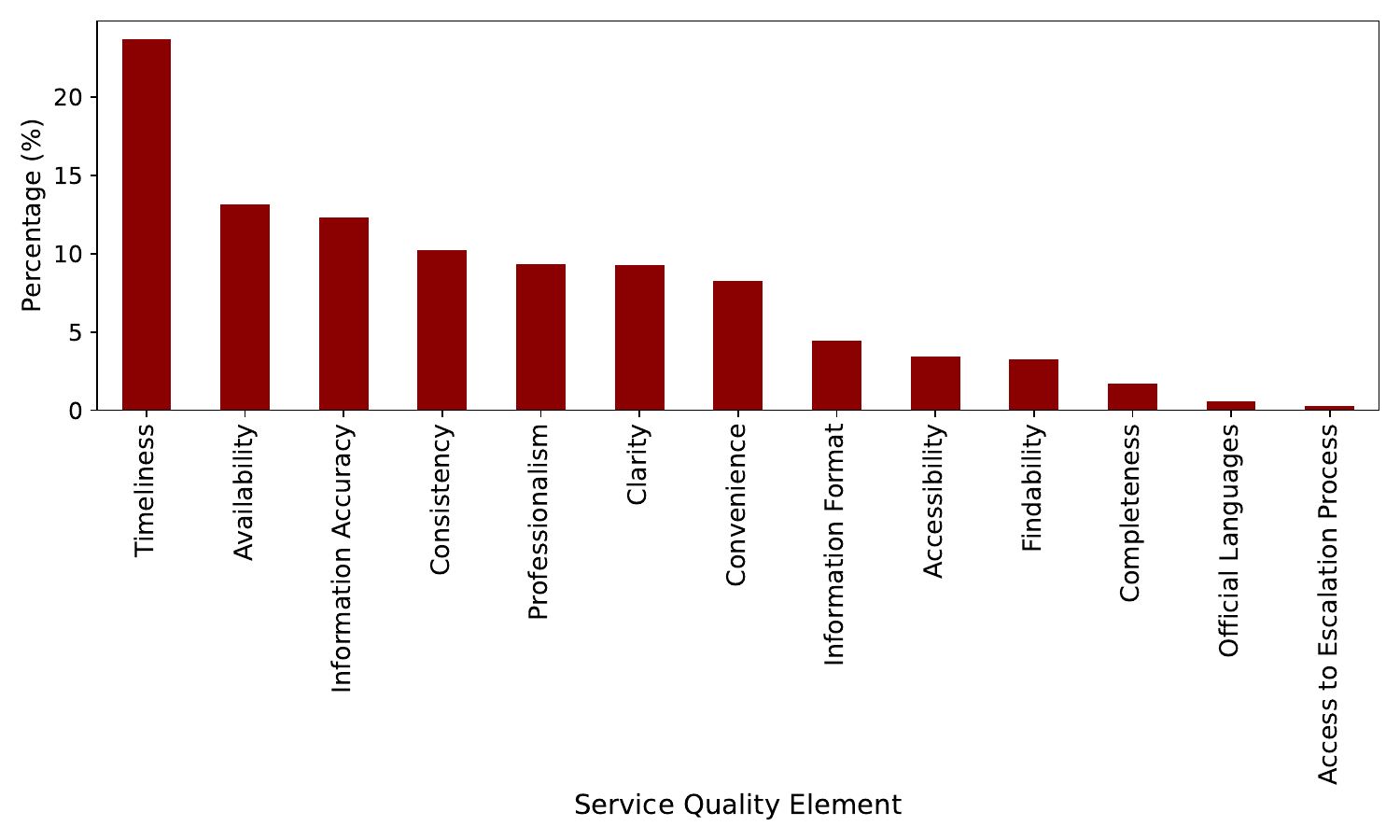}
\caption{Topic distribution across SQEs identified by the fine-tuned and quantized model (Mistral). The model closely mirrors the expert-labeled distribution, emphasizing  ``Timeliness'' while maintaining a balanced representation across other categories.}

\label{fig:sqe_m}
\end{figure}

In this section, we compare the topic categorization by Tax Service experts with those identified by our LLMs, including both a prompt engineering-based approach (Zephyr) and a fine-tuned quantized model (Mistral). As outlined earlier, the organization's experts played a critical role in validating the model output through a structured evaluation process. This human-in-the-loop step ensures that the automated classifications of the model align with the expert understanding of the Service Quality Elements and maintain the interpretability and trustworthiness of the feedback analysis system.
The distribution of the topics identified by the Tax Service experts, shown in Figure~\ref{fig:Topic_distribution}, is heavily skewed toward the category ``Timeliness''. This distribution likely reflects the prioritization of the timely response by experts, highlighting issues they deem most urgent within their capacity to assign only one Service Quality Element per feedback item.
In contrast, Figure~\ref{fig:sqe_z}, which shows the topic distribution of the prompt-engineering model, indicates a more balanced allocation among different SQEs. This suggests that the LLM captures a wider range of issues, offering a more diverse perspective on feedback topics. Although timeliness remains the most frequent category, its dominance is less pronounced compared to the human-assigned distribution, making the overall spread more comparable.
The concentration of human-labeled data on ``Timeliness'' may reflect the organizational emphasis on meeting predefined service standards within the Tax Service's Service Feedback Program, which prioritizes resolving taxpayer concerns within a specific timeframe. As a result, the reviewers may have focused more heavily on feedback related to response time and deadlines, aligning with operational goals. Although this focus ensures accountability in turnaround times, it may also inadvertently overshadow other, less obvious but still important service dimensions. 
In contrast, the broader distribution of the LLM output captures a wider variety of topics, including subtler concerns that may not be as immediately actionable, but are nevertheless critical to improving the overall taxpayer experience.
Finally, Figure~\ref{fig:sqe_m} shows how the fine-tuned and quantized model approaches topic categorization. Similarly to the expert collaborator distribution, this model also emphasizes ``Timeliness'', but does so in a way that maintains better balance across other service quality elements. This approach reflects a closer alignment with human review patterns, showing only a modest difference between ``Timeliness'' and other categories. By balancing attention between various elements, the fine-tuned model ensures that while timeliness is prioritized, other essential service quality dimensions are not overlooked, ultimately fostering a comprehensive and efficient approach to improving service quality.

%It is important to note that the range on the Y axis varies slightly between figures, reflecting the natural differences in model outputs and topic frequency distributions between SQEs. This variation is intentional and ensures that performance patterns are accurately represented. Furthermore, X-axis categories are ordered according to frequency within each model's output, allowing readers to more easily interpret which SQEs were prioritized the most frequently and how categorization patterns differ between models.

\subsubsection*{Similarity Measurement}

In this subsection, we evaluate the alignment between topics categorized by experts and those predicted by our models. To assess similarity, we matched topics assigned by the Tax Service experts to those generated by two variations of our Large Language Models.
The similarity score is calculated based on the proportion of unique topic groups where the expert-labeled Service Quality Element aligns with the predicted topics from the models. Instead of simply counting exact matches across the entire dataset, the measure considers matches within each group of related entries, making it more robust and representative of the true alignment.

\[
Similarity_{Topic} = \left( \frac{\text{Matched Feedback Instances}}{\text{Total Feedback Instances}} \right) \times 100
\]
where:\\
 \textbf{Matched Feedback Instances}
represents the number of individual feedback texts where the topic predicted by the model matches the expert labeled ``Service Quality Element''.
 \textbf{Total Feedback Instances}
is the total number of feedback texts in the dataset.

The comparison reveals a distinct difference in performance between the two models. The pretrained model achieves a lower similarity score of approximately 24.27\%, indicating limited alignment with expert-labeled topics at the individual feedback level. This lower score may reflect the broader generalization tendencies of the pre-trained model, which capture diverse topics but do not necessarily align with the specific categorizations made by experts. On the other hand, the fine-tuned quantized model achieves a higher similarity score of a much higher 66.64\%. This improvement suggests that the model, after fine-tuning and quantization, becomes more adept at recognizing and aligning with the feedback topics deemed important by the experts. The fine-tuning process likely helps the model focus on the nuanced context-specific themes and language present in the feedback texts, leading to a better alignment with expert judgment. 
In summary, while the pre-trained model offers a wider scope of topic identification, the fine-tuned quantized model demonstrates a stronger capability to reflect the expert-labeled categorizations at the individual feedback level, resulting in higher similarity scores and a more accurate representation of expert-defined topics.

\subsubsection*{Expert Evaluation of Model Applicability}

In addition to measuring similarity between model outputs and Tax Service-labeled SQEs, we conducted a survey with Tax Service Feedback Officers to assess the practical applicability of the model with the lowest similarity score. This evaluation offered a complementary perspective based on expert judgment rather than strict label matching. While similarity scores provide a quantitative benchmark, they may not fully capture the model’s real-world utility, given the subjective nature of language and topic classification.
To address this, we implemented a human-in-the-loop evaluation using bilingual feedback samples randomly selected to avoid bias. The model assigned SQEs to each sample, and experts were asked to review the assignments, indicate agreement or disagreement, and propose alternatives where necessary.
We then conducted pairwise t-tests on 51 SQE instances derived from 10 randomly selected feedback texts (in both English and French). This sample size satisfies the Central Limit Theorem's requirement (n > 30), allowing for valid inference. The goal was to test whether differences between model and expert categorizations were statistically significant.
Results, summarized in Table~\ref{tab:Fcompare}, showed no significant differences at the 95\% confidence level. This indicates that the model’s classifications are statistically indistinguishable from those of Tax Service experts. Moreover, alignment was observed across all five reviewers, suggesting that the model effectively mirrors human scoring patterns.

\begin{table}[h!]
    \centering
    \renewcommand{\arraystretch}{1.2} % Adjust row spacing for readability
    \begin{tabular}{c c c c}
        \toprule
        \textbf{Comparison} & \textbf{T-value} & \textbf{P-value} & \textbf{Significance} \\
        \midrule
        Model vs. Officers & -1.968 & 0.085 & No significant difference \\
        \bottomrule
    \end{tabular}
    \caption{T-test results comparing the model’s scores against the most divergent expert evaluator}
    \label{tab:Fcompare}
\end{table}

Additional comparisons among Service Feedback Program evaluators yielded similar results, without significant differences. These findings support the conclusion that the LLM scoring is comparable to the expert scoring, with no statistically significant deviation in the comparisons. The close alignment between the model and the expert evaluators suggests that the model can reliably approximate human assessment, making it a valuable tool for scaling evaluations and maintaining accuracy in feedback analysis.
This evaluation, which combines similarity measures with subjective expert judgment and statistical validation, ensures that the LLM is not only aligned with expert categorizations but also applicable and effective in practical, user-centered contexts.

\subsection{Trend Detection Result}

As described in the methodology, trend analysis begins with segmenting the dataset into two time periods using a flexible quantile cutoff. A cutoff of 0.5, for instance, splits the data into two 6-month intervals, while other values can isolate specific events or seasonal effects. This segmentation enables targeted analysis of temporal patterns. 
Categorical feedback topics were then label-encoded for modeling. We trained separate multinomial logistic regression models for each period, fitting 26 binary regressions—two for each of the 13 SQE categories. The sample size per model reflected the frequency of each topic label, as shown in Figure~\ref{fig:sqe_m}.
To detect shifts over time, we compared categorical variable coefficients across periods using 95\% bootstrap confidence intervals. Trends were classified as emerging, persistent, or disappearing based on coefficient changes. An illustrative example (Table~\ref{tab:example}) shows how logistic regression identifies temporal topic dynamics while respecting data privacy.

\begin{table}[h]
    \centering
    \begin{tabular}{lll}
        \toprule
        \textbf{Service Quality Element} & \textbf{Feature} & \textbf{Trend} \\
        \midrule
        ACCESSIBILITY & Age $\leq$ 19 & Disappearing \\
        CLARITY & Age $\geq$ 60 & Emerging \\
        CLARITY & 19 $\leq$ Age $\leq$ 60 & Emerging \\
        TIMELINESS & Gender: F & Emerging \\
        TIMELINESS & Gender: M & Emerging \\
        TIMELINESS & Preferred Language: English & Emerging \\
        \bottomrule
    \end{tabular}
    \caption{Illustrative Example of Trend Detection Based on Service Quality Elements}
    \label{tab:example}
\end{table}
The table highlights several notable trends in feedback patterns over time. One key observation is a disappearing trend in accessibility concerns among users under 19. In the second period, reports of accessibility issues from this group decreased, suggesting either an improvement in services or a reduced perception of barriers. Using this tool, the organization can now identify whether specific measures taken to improve the taxpayer experience have become effective.

In contrast, emerging trends indicate that clarity issues have become more prominent among older adults (60 and older) as well as the middle-aged group (19 to 60 years). This shift suggests a growing concern or a change in how these age groups perceive communication and information delivery.
A similar upward trend is observed in timeliness-related feedback, which has increased among both male and female users, as well as those who prefer English. This pattern may point to a broader issue with service delays or responsiveness, which requires further attention.
These insights illustrate how logistic regression modeling and coefficient comparison can reveal meaningful shifts in user feedback. By identifying disappearing, emerging, and persistent trends, this approach supports data-driven decision making and helps improve service delivery.

\section{Discussion}

This study introduced a multilingual human-in-the-loop system that integrates fine-tuned and quantized LLMs with statistical trend detection to categorize taxpayer feedback into service quality elements and uncover evolving concerns across demographic groups. By combining advanced natural language processing, resource-efficient model deployment, and expert validation, the framework not only enhances the scalability of public service feedback analysis but also promotes equity, transparency, and operational alignment in government service delivery. The model demonstrates alignment with expert-labeled categorizations and effectively identifies emerging, persistent, and disappearing concerns in multilingual feedback.

\subsection{Limitations}
Despite the promising results of this study, there are some limitations that must be recognized. First, and due to system availability, the dataset used for analysis was limited to a single year of feedback, which may not fully capture long-term or seasonal trends.
Finally, the models used, although fine-tuned, were constrained by the available computational resources, limiting the extent of fine-tuning and the depth of analysis that could be achieved by using more powerful models (36B parameter+). These constraints may affect the generalizability and scalability of the findings across different contexts or larger datasets.

\subsection{Future Work}
Future research should focus on addressing these limitations by expanding the scope of the analysis to include more extensive multi-year datasets. Incorporating feedback in additional languages could enhance the inclusivity and accuracy of the system, providing a more comprehensive understanding of diverse taxpayer experiences. Also, future work could explore the use of advanced non-linear modeling techniques such as neural network-based classifiers or ensemble learning approaches, which may offer improved performance in trend detection. Integrating continuous learning mechanisms, where the model updates and refines its analysis in real-time, could further enhance adaptability and responsiveness. Furthermore, implementing a more interactive human-in-the-loop system could provide ongoing validation and refinement of the model output, ensuring alignment with real-world expectations and needs. By addressing these areas, the proposed framework can evolve into a robust tool for dynamic, multilingual feedback analysis, and trend monitoring at scale.

\subsection{Theoretical Implications}

This research contributes to the growing literature on AI-assisted public service analysis by demonstrating how domain-specific fine-tuning and quantization of LLMs can be systematically applied to detect trends in structured feedback. Unlike traditional topic modeling approaches that prioritize unsupervised or general-purpose tasks, this study operationalizes topic modeling based on predefined expert taxonomies. This bridges a key theoretical gap between thematic modeling and institutional interpretability.
Furthermore, the integration of stratified logistic regression with bootstrapped confidence intervals introduces a novel fairness-aware approach for detecting demographic disparities in topic prevalence. This supports broader theoretical efforts in explainable and equitable NLP, showing how statistical reasoning can complement LLM predictions for rigorous, demographically contextualized trend analysis.
The research also contributes to the theory of human-AI teaming by embedding expert validation at multiple stages—preprocessing, model fine-tuning, and output review. This addresses confabulation risk in LLMs and responds to ongoing academic calls for ethically aligned, human-supervised AI in sensitive domains like taxation.

\subsection{Practical Implications}

The practical contributions of this study lie in its ability to translate advanced AI techniques into actionable tools for public service institutions. By developing a multilingual feedback analysis system that incorporates fine-tuned and quantized LLMs, the research offers a scalable solution for organizations to manage increasing volumes of unstructured feedback efficiently. This approach reduces the dependency on manual review processes, allowing institutions to process and categorize service feedback with greater speed and consistency.
Importantly, the framework is designed with resource constraints in mind. The quantization of the LLM significantly lowers the computational burden, making it feasible to deploy the system even within infrastructure-limited environments. This ensures that smaller agencies or departments, which may lack access to high-end servers or GPUs, can still benefit from the capabilities of modern NLP tools.
Beyond technical efficiency, the system enhances the quality of public service by enabling real-time identification of emerging concerns across different demographic groups. By analyzing topic trends through demographic lenses—such as age, language, and gender—the system allows decision-makers to detect systemic disparities that might otherwise go unnoticed. This supports more equitable service delivery by highlighting the specific needs of underserved populations and enabling timely organizational response.
Moreover, the integration of explainable outputs and expert validation strengthens trust in the system's recommendations. The ability to trace the model’s rationale for each topic assignment ensures transparency and allows human reviewers to engage meaningfully with the results. Overall, the framework provides a practical path toward responsible AI adoption in the public sector, offering both operational improvements and enhanced fairness in citizen engagement.

\section{Conclusions}

In this research, we presented a unique approach to analysis of service feedback data that integrates state-of-the-art techniques for identifying feedback patterns, aimed at reducing biases and enhancing service quality. 
Our methodology combines statistical modeling with natural language processing techniques, integrating state-of-the-art large language models for topic modeling with traditional techniques for trend analysis, while also incorporating human-in-the-loop methods that incorporate the subject matter expertise of Service Feedback officers. This hybrid approach ensures that the system is not only effective and scalable to detect trends within multilingual feedback data, but also contextually grounded and aligned with operational realities.
We also explored how fine-tuning and quantization of the LLM results in alignment with the specific idiosynchratic evaluation processes within the organization, while simultaneously optimizing resource usage. Our suggestion is that whenever an LLM is considered to be deployed within an organization that has developed its own culture and language, this fine-tuning is performed. Otherwise, the model will most likely fail to understand these nuances and miss important details, which will translate into reduced performance. Our results demonstrate that the fine-tuned model, customized with Tax Service-specific text data, was more closely aligned with expert opinion in terms of the categorization of topics. This alignment was evident in the similarity analysis and was further supported by an evaluation survey, where the Tax Service experts ranked the models according to their output. The fine-tuned model showed significant improvements, capturing topics and nuances that better reflected expert assessments.
The second stage of our proposal involved the use of these nuanced outputs for trend analysis. Our findings indicate that our methodology effectively identified emerging, persistent, and disappearing trends in diverse demographic groups, focusing on specific themes that require targeted improvements. By employing a multiphase approach, including text tokenization, de-identification, model fine-tuning, and trend detection, we ensured a thorough analysis of feedback data within specific demographics. Our framework adheres to key principles of fairness, transparency, explainability, and accountability, providing a powerful tool to improve the quality of feedback services while addressing potential biases in feedback analysis.

%\printbibliography
\bibliographystyle{elsarticle-num}
\bibliography{ref.bib}

\FloatBarrier % Ensures table stays in Appendix C

\end{document}